\documentclass[runningheads]{llncs}

\usepackage{eccv}

\usepackage{eccvabbrv}

\usepackage{graphicx}
\usepackage{booktabs}

\usepackage[accsupp]{axessibility}  

\usepackage{url}
\usepackage{algorithm}
\usepackage{algpseudocode}
\usepackage{pifont}
\usepackage{xcolor}
\usepackage{multirow}
\usepackage{tabularx}
\usepackage{booktabs}

\usepackage{xspace}
\usepackage{amsmath}
\usepackage{amssymb}
\usepackage[table,xcdraw]{xcolor}
\usepackage{multirow}
\usepackage{makecell}
\usepackage{pifont}
\usepackage{arydshln}

\usepackage{todonotes}
\usepackage{fontawesome}
\usepackage{bm}
\usepackage{booktabs}
\usepackage[colorlinks=true, linkcolor=blue, citecolor=blue, urlcolor=blue]{hyperref}
\usepackage{wrapfig}

\usepackage{listings}
\usepackage{xcolor}
\usepackage[most]{tcolorbox}

\usepackage{array}
\usepackage{graphicx}
\usepackage{soul}
\usepackage{tcolorbox}
\usepackage{listings}
\usepackage{xcolor}

\usepackage[numbers, sort&compress]{natbib}
\usepackage{wrapfig}

\newcommand{\ourmethod}{Skin-R1}
\definecolor{prompttitle}{HTML}{003366} 
\definecolor{lightblueborder}{HTML}{BFD7FF} 

\usepackage[dvipsnames]{xcolor}
\usepackage[most]{tcolorbox}
\usepackage{xcolor}
\definecolor{darkgreen}{RGB}{0,100,0}   
\definecolor{skyblue}{RGB}{135,206,235} 

\newtcolorbox{promptbox}[2][]{%
  colback=white,
  colframe=lightblueborder,
  coltitle=prompttitle,
  title=\textbf{#2},
  fonttitle=\small\bfseries,
  boxrule=0.5pt,
  arc=3pt,
  outer arc=3pt,
  top=4pt, bottom=4pt, left=6pt, right=6pt,
  breakable,
  listing only,
  verbatim
}
\algrenewcommand{\algorithmicrequire}{\textbf{Input:}}
\algrenewcommand{\algorithmicensure}{\textbf{Output:}}

\usepackage{hyperref}

\usepackage{orcidlink}

\newcommand{\blfootnote}[1]{%
  \begingroup
  \renewcommand\thefootnote{}\footnote{#1}%
  \addtocounter{footnote}{-1}%
  \endgroup
}

\begin{document}

\title{Skin-R1: Clinical Knowledge-Guided Dermatological Diagnosis Using Vision-Language Models}

\titlerunning{Skin-R1}




\author{Zehao Liu\inst{1} \and
Weijieying Ren\inst{2} \and
Jipeng Zhang\inst{3} \and
Tianxiang Zhao\inst{1} \and
Jingxi Zhu\inst{1} \and
Xiaoting Li\inst{1} \and
Vasant G Honavar\inst{1}}

\authorrunning{Z. Liu et al.}


\institute{Pennsylvania State University, University Park, PA 16802, USA\\
\email{\{zml5418,jqz5678,vuh14\}@psu.edu},
\email{\{txiangchao,xiaotili0123\}@gmail.com} \and
Stanford University, Stanford, CA 94305, USA\\
\email{wjyren@stanford.edu} \and
Hong Kong University of Science and Technology, Hong Kong, China\\
\email{jzhanggr@connect.ust.hk}}

\maketitle
\blfootnote{Accepted to ECCV 2026.}
\begin{abstract}
Vision--language models (VLMs) have recently shown promise for assisting clinical reasoning in dermatological diagnosis. However, their trustworthiness and clinical utility remain limited by three key challenges: heterogeneous datasets with inconsistent diagnostic labels and concept annotations, the lack of grounded diagnostic rationales for reliable reasoning supervision, and limited scalability when transferring knowledge from small, densely annotated datasets to large collections with sparse labels.

To address these challenges, we propose \ourmethod{}, a dermatology-oriented VLM that integrates textbook-grounded clinical reasoning supervision with reinforcement learning (RL) to improve the accuracy and robustness of diagnostic prediction.
First, we construct a textbook-based reasoning generator that synthesizes hierarchy-aware and differential-diagnosis (DDx) diagnostic trajectories derived from authoritative dermatology knowledge. 
Second, these trajectories are used for supervised fine-tuning (SFT), establishing a clinically grounded reasoning foundation for the model. 
Finally, we introduce an RL training framework that incorporates the hierarchical structure of dermatological diseases into the reward design, enabling the model to generalize grounded diagnostic reasoning to large-scale datasets with sparse annotations.

Extensive experiments across multiple dermatology benchmarks demonstrate that \ourmethod{} consistently improves diagnostic accuracy and robustness compared to state-of-the-art Med-VLM baselines. Ablation studies further highlight the critical role of grounded reasoning supervision introduced during the SFT stage.

Code and model weight are available at \url{https://github.com/l593191569/Skin-R1}.

\end{abstract}

\begin{table*}[!t]
\centering
\caption{Comparison of representative Medical VLMs in terms of key diagnostic reasoning abilities. 
\textbf{Grounded Rationale} indicates whether the model is designed to produce rationales grounded in authoritative clinical sources;
\textbf{DDx Awareness} reflects awareness of differential diagnoses through comparative reasoning; 
\textbf{Hierarchical Awareness} captures understanding of disease taxonomy and multi-level relations; 
and \textbf{Sparse-Supervision Learning} measures the ability to learn and generalize from sparsely annotated data.  }
\label{tab:comparison_training_recipes}
\resizebox{\textwidth}{!}{
\begin{tabular}{c|l|cccc}
\toprule
\rowcolor[HTML]{E9F3FE}
\textbf{Type} & \textbf{Model} & \textbf{Grounded Rationale}& \textbf{DDx Awareness}   & \textbf{Hierarchical Awareness} & \textbf{Sparse-Supervision Learning} \\
\midrule \midrule

\multirow{5}{*}{\textbf{RL-based}} 
& Med-R1~\citep{lai2025med} & \ding{55} & \ding{55}  & \ding{55} & \ding{51} \\
& MedVLM-R1~\citep{pan2025medvlm} & \ding{55} & \ding{55}  & \ding{55}& \ding{51} \\
& MedCCO~\citep{rui2025improving} & \ding{55} & \ding{55} & \ding{55}  & \ding{51} \\
& RARL~\citep{pham2025rarl} & RL reward & \ding{55} & \ding{55} & \ding{51} \\
& \citep{zhu2025toward} & RL reward & \ding{55} & \ding{55}& \ding{51}  \\

\midrule
\multirow{4}{*}{\textbf{SFT-based}} 
& SkinVL-PubMM~\citep{zeng2025mm} & grounding SFT & \ding{55}  & \ding{55} & \ding{55} \\
& OmniV-Med~\citep{jiang2025omniv} & grounding SFT & \ding{51} & \ding{55} & \ding{55} \\
& LLaVA-Med~\citep{li2023llava} & distillation & \ding{51}  & \ding{55} & \ding{55} \\
& HuatuoGPT-Vision~\citep{chen2024huatuogpt} & distillation & \ding{51} & \ding{55} & \ding{55} \\

\midrule
\multirow{2}{*}{\textbf{Hybrid}} 
& MedGemma~\citep{sellergren2025medgemma} & grounding SFT, distillation, RL reward & \ding{51}  & \ding{55} & \ding{51} \\
& \cellcolor[HTML]{DAE0FB}\textbf{\ourmethod} & \cellcolor[HTML]{DAE0FB}\textbf{grounding SFT} & \cellcolor[HTML]{DAE0FB}\textbf{\ding{51}}  & \cellcolor[HTML]{DAE0FB}\textbf{\ding{51}} & \cellcolor[HTML]{DAE0FB}\textbf{\ding{51}} \\

\bottomrule
\end{tabular}
}
\end{table*}

\section{Introduction}
\label{Introduction}

Dermatological diagnosis is fundamentally a visual reasoning task: clinicians interpret subtle visual patterns, compare them against hierarchical disease taxonomies, and perform differential diagnosis (DDx) to distinguish visually similar conditions. Recent advances in multimodal vision--language models (VLMs) suggest that AI systems could assist this process by jointly interpreting medical images and clinical knowledge. However, enabling such systems to perform reliable and scalable diagnostic reasoning remains an open challenge.

Each year, millions of people worldwide experience delayed or incorrect diagnosis of skin conditions such as melanoma, eczema, and psoriasis—often due to limited access to dermatologists \citep{hay2014global,feng2018comparison,resneck2004dermatology} or subjective interpretation of visual symptoms. Advances in computer vision, deep learning, and multimodal learning now offer the possibility of AI-assisted systems capable of analyzing dermoscopic and clinical images with expert-level precision, supporting clinicians in triage, screening, and treatment planning. By combining clinical expertise with AI, dermatology-oriented vision models have the potential to democratize access to high-quality care and reduce the global burden of preventable skin disease \citep{zhou2024pre,liu2020deep,cruz2013deep,yuan2017automatic}.

Recent multimodal vision foundation models and medical vision--language models (Med-VLMs) \citep{yan2025make,kim2024transparent,yan2025multimodal,koh2020concept,yan2023towards,hu2024towards} have demonstrated promising performance in dermatological image understanding and diagnosis. However, existing approaches still face important limitations. For example, models such as SkinGPT-4 \citep{zhou2024pre} rely heavily on labor-intensive human annotations and often lack clinically grounded diagnostic rationales.

Building a reliable dermatology vision--language model (VLM) presents three major challenges.

\begin{enumerate}
\item \textbf{Data heterogeneity.}
Real-world dermatology datasets vary widely in their diagnostic ontologies and concept vocabularies. For example, SkinCon~\citep{daneshjou2022skincon} provides annotations for 29 dermatological concepts (e.g., \emph{erythema}, \emph{plaque}) but only two coarse diagnostic labels (benign vs.\ malignant). In contrast, DermNet~\citep{dermnet2024} contains images spanning more than 600 diseases but lacks structured concept annotations. Such discrepancies lead to inconsistent supervision and noisy training signals.

\item \textbf{Lack of expert-like reasoning supervision.}
Accurate dermatological diagnosis requires structured reasoning that respects hierarchical disease taxonomies and supports differential diagnosis (DDx). However, existing datasets rarely provide supervision that captures clinically grounded diagnostic reasoning processes.

\item \textbf{Limited scalability of training paradigms.}
Models trained on small datasets with dense annotations often struggle to generalize to large datasets with sparse labels, making it difficult to transfer grounded diagnostic reasoning to large-scale heterogeneous data.
\end{enumerate}

These challenges reflect a deeper issue: dermatological diagnosis is fundamentally a reasoning problem over a hierarchical disease space, where visually similar conditions must be distinguished through structured clinical reasoning and differential diagnosis. Existing training paradigms struggle to capture such reasoning processes while remaining scalable to large heterogeneous datasets.

Motivated by this observation, a key open question is how to develop dermatology VLMs capable of performing \emph{expert-like clinical reasoning} while remaining scalable to large heterogeneous datasets. Existing approaches typically rely either on small datasets with dense annotations that provide limited scale, or on large weakly labeled datasets that lack reliable reasoning supervision. Bridging this gap requires a framework that can first acquire grounded clinical reasoning patterns from reliable sources and then effectively generalize them across diverse datasets.

\begin{figure*}[h]
\centering
\includegraphics[width=\textwidth]{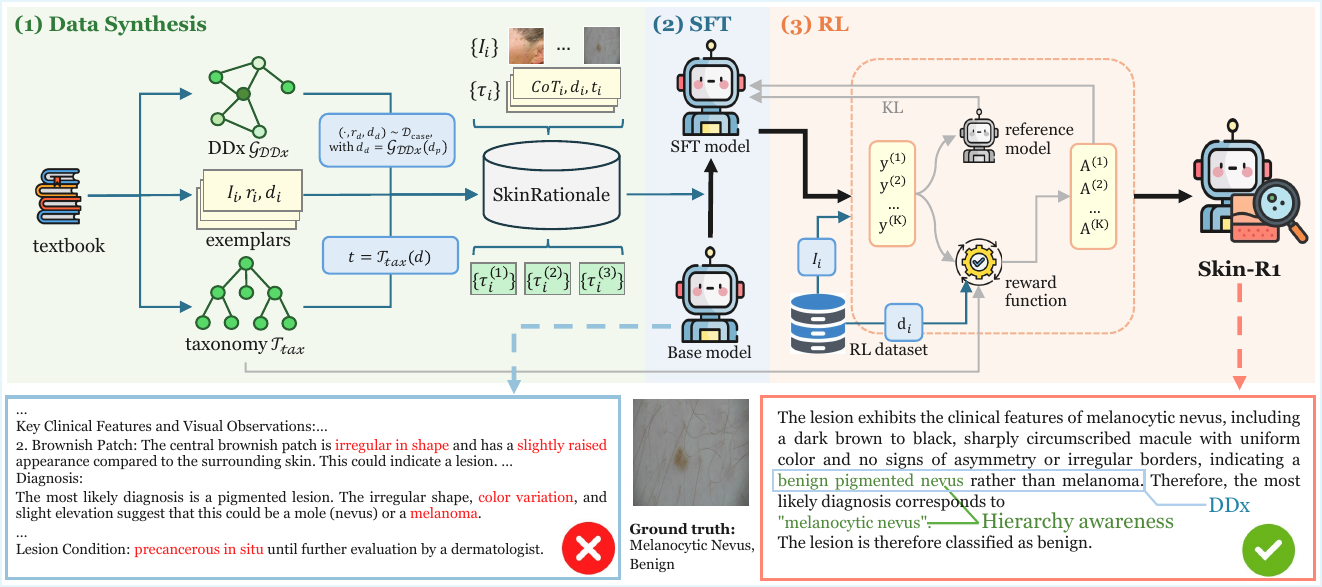}
\caption{
\textbf{Top:} Overview of the proposed training framework, illustrating 
(1) the synthesis of the SkinRationale dataset, 
(2) the SFT stage performed on SkinRationale, and 
(3) the GRPO-based RL stage. \\
\textbf{Bottom:} Two representative cases comparing diagnostic responses from \ourmethod{} and baseline models. The response from \ourmethod{} is concise, accurate, and aligned with expert diagnostic reasoning. 
\textcolor{red}{Red text} highlights incorrect or hallucinated content; \textcolor{darkgreen}{Green text} denotes \emph{hierarchical-aware} reasoning; and content enclosed within the \textcolor{skyblue}{\textbf{blue box}} indicates the \emph{DDx comparison} derived from prior differential diagnostic reasoning.}
\label{fig:framework}
\end{figure*}

To address this problem, we propose \ourmethod{}, which, to the best of our knowledge, is the first dermatology-oriented VLM that integrates \emph{textbook-grounded clinical reasoning} with reinforcement learning--based adaptation. Our framework unifies expert-aligned reasoning and scalable knowledge transfer within a single end-to-end training pipeline. Figure~\ref{fig:framework} illustrates the overall architecture and training procedure of \ourmethod{}.

First, we construct a grounded dataset of hierarchy-aware and differential-diagnosis (DDx)-informed diagnostic trajectories derived from authoritative dermatology textbooks. We then perform supervised fine-tuning (SFT) on these trajectories to establish a clinically grounded reasoning foundation. Finally, we scale this reasoning capability to large datasets with sparse annotations through a reinforcement learning (RL) framework guided by hierarchical diagnostic accuracy and structured output regularization rewards, encouraging the model to generalize grounded diagnostic reasoning across heterogeneous data sources.

The main contributions of this paper are summarized as follows:

\begin{enumerate}

\item We introduce \ourmethod{}, a novel training paradigm for dermatology vision--language models that integrates textbook-grounded clinical reasoning with reinforcement learning to improve diagnostic accuracy and robustness across heterogeneous datasets.

\item We design a textbook-grounded trajectory generation framework that synthesizes hierarchy-aware and differential-diagnosis (DDx)-informed diagnostic reasoning traces, providing scalable supervision for clinically consistent reasoning.

\item We develop a hierarchical reward design that incorporates disease taxonomy structure into reinforcement learning, enabling grounded reasoning patterns learned from dense data to generalize effectively to large-scale sparsely annotated datasets.

\item We conduct extensive experiments across multiple dermatology benchmarks demonstrating that \ourmethod{} consistently outperforms state-of-the-art baselines in diagnostic accuracy and robustness.

\end{enumerate}

More broadly, our results suggest that grounding vision-language models in expert-level reasoning supervision may provide a promising path toward increasing the reliability of AI systems for medical applications.

\section{Related Work}
\label{sec:related}
\paragraph{\textbf{Medical Vision--Language Models.}}

Vision--language models (VLMs)~\citep{bai2025qwen2,liu2023llava,zhu2023minigpt,yin2023survey,awais2025foundation} extend large language models (LLMs)~\citep{gpt4,touvron2023llama,team2023gemini} to multimodal reasoning by enabling joint processing of visual and textual inputs.
Medical VLMs (Med-VLMs)~\citep{ren2025comprehensive,zhou2023survey,tian2024role} adapt these architectures for clinical applications by developing domain-specific visual encoders, integrating them with LLMs, and aligning multimodal representations across diverse biomedical datasets.

Early efforts focused on improving visual understanding in specialized medical domains through medical foundation models, including radiology~\citep{wang2022medclip}, pathology~\citep{lu2024visual}, and dermatology~\citep{yan2025make,kim2024transparent,yan2025multimodal,yan2026vision}. 
Subsequent approaches introduced encoder--decoder Med-VLM architectures that combine medical visual encoders with language models. 
For example, XrayGPT~\citep{thawkar2023xraygpt} integrates MedCLIP~\citep{wang2022medclip} with a fine-tuned LLM, while LLaVA-Med~\citep{li2023llava} improves multimodal alignment using PubMed image--text pairs. 
BiomedGPT~\citep{luo2024biomedgpt} adopts a BERT--GPT hybrid architecture trained on biomedical corpora, and HuatuoGPT-Vision~\citep{chen2024huatuogpt} scales this paradigm using the large PubMedVision dataset. 
More recently, HealthGPT~\citep{lin2025healthgpt} unifies medical image understanding and generation within a single autoregressive framework, and SkinGPT-4~\citep{zhou2024pre} applies instruction tuning to improve dermatological reasoning.

Despite these advances, most Med-VLMs rely on large volumes of high-quality annotated data, limiting their applicability in domains where expert annotations are scarce. 
In contrast, \ourmethod{} learns grounded diagnostic reasoning from a small set of high-quality expert-derived trajectories and then scales this capability using reinforcement learning.

\paragraph{\textbf{RL for Reasoning in Med-VLMs.}}

Recent work has demonstrated that reinforcement learning (RL) can induce reasoning abilities in large language and vision--language models~\citep{guo2025deepseek,xie2025logic,chen2025enigmata,hu2025beyond,zhang2025r1}. 
Methods based on RL with verifiable rewards (RLVR), such as DeepSeek-R1~\citep{guo2025deepseek}, employ Group Relative Policy Optimization (GRPO) to encourage structured reasoning behaviors without explicit chain-of-thought (CoT) supervision.

RL has also been applied to improve reasoning in Med-VLMs for diagnostic tasks~\citep{lai2025med,pan2025medvlm,rui2025improving,pham2025rarl,zhu2025toward,liu2026skinflow}. 
Because high-quality annotated medical reasoning data are scarce, many of these approaches bypass supervised fine-tuning (SFT) and instead apply RL directly using GRPO-style objectives. 
This RL-only paradigm has shown promising improvements in medical visual question answering and cross-domain generalization.

For example, MedVLM-R1~\citep{pan2025medvlm} demonstrates that GRPO can induce emergent chain-of-thought reasoning without explicit reasoning supervision. 
Similarly, Med-R1~\citep{lai2025med} reports that models trained without explicit reasoning supervision can sometimes outperform those trained with CoT-style supervision, suggesting that conventional SFT may introduce \emph{memorization shortcuts}. 
However, we argue that this phenomenon arises because existing approaches lack clinically grounded reasoning supervision, often leading to heuristic or hallucinated rationales~\citep{lanham2023measuring,huanglarge}.

Unlike prior work, \ourmethod{} first establishes a grounded diagnostic reasoning foundation using textbook-derived diagnostic trajectories and then employs RL to amplify and generalize this reasoning capability. 
Ablation experiments (\cref{exp:ablation}) confirm that this combination of grounded supervision and RL significantly improves diagnostic reasoning performance.
\section{Methods}
\label{sec:method}

\paragraph{Pipeline Overview.}
The top panel of Figure~\ref{fig:framework} illustrates the overall training pipeline of \ourmethod{}. 
Our framework consists of three stages. 
First, we construct \textbf{SkinRationale}, a textbook-derived dataset that captures hierarchy-aware and differential diagnosis (DDx)-informed reasoning trajectories grounded in authoritative dermatology knowledge. 
Second, we perform supervised fine-tuning (SFT) on these trajectories to initialize the model with clinically consistent diagnostic reasoning. 
Finally, we apply reinforcement learning (RL) using Group-Relative Policy Optimization (GRPO) to generalize this reasoning ability to large-scale dermatology datasets with sparse annotations. 
Through this pipeline, the model learns to produce structured diagnostic reasoning while maintaining robustness across heterogeneous clinical data distributions.

In the following sections, we first describe the construction of SkinRationale,
and then present the two-stage training strategy consisting of SFT and GRPO.
\subsection{Training Data for Grounded Reasoning}

Reliable dermatological diagnosis requires resources that capture multi-step clinical reasoning. However, existing dermatology datasets primarily focus on visual classification or lesion segmentation and rarely include diagnostic rationales linking visual concepts to clinical conclusions. To address this limitation, we introduce \textbf{SkinRationale}, a textbook-derived dataset designed to provide clinically grounded reasoning supervision.

\subsubsection{Extracting Textbook-Grounded Knowledge}

We extract three complementary components from the dermatology textbook of~\citep{kang2019fitzpatrick}:
(1) Diagnostic exemplars $\mathcal{D}= \{(I_i, r_i, d_i)\}_{i=1}^{N}$, each represented as a triplet $(I_i, r_i, d_i)$ consisting of a clinical image $I_i$, a textual rationale $r_i$, and a diagnostic label $d_i$,
(2) A differential diagnosis graph $\mathcal{G}_{\text{DDx}}$, where edges connect clinically confusable diseases; and
(3) A disease taxonomy $\mathcal{T}_{\text{tax}}$, where parent–child relations represent hierarchical disease categories.

In total, we collect 220 diagnostic exemplars, construct a DDx graph with 211 nodes and 245 edges (covering 61.36\% of the collected exemplars), and build a disease taxonomy containing 458 nodes and 473 edges.

The knowledge extraction pipeline largely follows the automated procedure proposed in~\citep{kim2024transparent}. The pipeline consists of five stages: 
(1) image and text extraction, 
(2) image clustering and filtering, 
(3) text pairing and filtering, 
(4) diagnostic rule extraction, and 
(5) DDx and taxonomy confirmation and refinement. 
Human involvement is limited to selecting representative image clusters. Further implementation details are provided in Appendix~\cref{appendix:data_extraction}.

\subsubsection{Constructing Reasoning Trajectories}

Generating reliable reasoning trajectories for cold-start supervision is challenging. Prior work often relies on large reasoning models (LRMs) such as QwQ-32B~\cite{qwq32b} to produce reasoning traces. However, such model-generated rationales may contain partial errors, potentially leading to model collapse~\citep{shumailov2024ai} or hallucination propagation~\citep{wang2025comprehensive}. 

Instead, we construct reasoning trajectories from textbook-derived rationales, which provide stable and expert-curated supervision signals.

Two structural challenges arise when converting textbook knowledge into training trajectories: 
\begin{enumerate}
\item \textbf{Label granularity inconsistency}: diagnoses may appear at different specificity levels (e.g., melanoma vs.\ superficial spreading melanoma); and \item \textbf{Differential diagnosis ambiguity}: dermatological reasoning requires distinguishing visually similar diseases.
\end{enumerate}
To address these issues, we introduce two complementary mechanisms.

\paragraph{Hierarchical diagnosis completion.}

We construct a dermatology diagnosis tree capturing parent–child relationships among disease entities (e.g., melanoma $\rightarrow$ superficial spreading melanoma). Each final label is augmented with its ancestral and sibling nodes. This process mitigates inconsistencies in label granularity and encourages taxonomic coherence across diagnostic levels.

\paragraph{Differential diagnosis reasoning enrichment.}

Using the DDx graph, we synthesize reasoning trajectories that explicitly describe how visually similar conditions are differentiated. These DDx-informed rationales allow the model not only to identify the correct diagnosis but also to provide interpretable justification for excluding alternatives.

The resulting reasoning trajectories are defined as

\begin{equation}
\{\tau_i\}_{i=1}^{K} = \{(CoT_i, d_i, t_i)\}_{i=1}^{K},
\end{equation}

where each trajectory contains a reasoning trace $CoT_i$, a diagnosis label $d_i$, and hierarchical diagnostic metadata $t_i$. In total, we construct $K = 2020$ diagnostic trajectories. Details of trajectory synthesis are provided in Appendix~\cref{appendix:dataset_construction}.

\subsection{Enhancing Grounded Medical Reasoning}

\paragraph{Problem Formulation.}
We formulate dermatological diagnosis as a multimodal reasoning task. Given a clinical image $I$ and an instruction prompt $p$, the model generates a structured diagnostic response consisting of: (i) a reasoning trajectory $CoT$, (ii) a predicted disease label $\hat{\ell}$, and (iii) a lesion category $\hat{b}$ (e.g., benign, malignant, or precancerous). Formally, the policy model $\pi_\theta$ learns a conditional distribution 
\begin{equation}
    \pi_\theta(y \mid I, p),
\end{equation}
 where $y = (CoT, \hat{\ell}, \hat{b})$ denotes the structured output. The objective is to learn a policy that produces diagnostically correct predictions while maintaining clinically coherent reasoning trajectories.

\subsubsection{Supervised Fine-Tuning on SkinRationale}

We first train the policy model $\pi_{\theta}$ using SFT on hierarchy-aware and DDx-informed diagnostic trajectories. Given a clinical image $I$, an instruction prompt $p$, and the trajectory
$
\tau = (x_0, x_1, \ldots, x_{n}),
$
the training objective is

\begin{equation}
L = - \sum_{i=1}^{n} \log \pi_{\theta}(x_i \mid x_{<i}, I, p).
\end{equation}

This objective allows the model to imitate expert diagnostic reasoning across hierarchical disease categories and establishes a robust initialization for RL optimization.

\subsubsection{Reinforcement Learning via GRPO}

Although SFT instills grounded reasoning ability, it remains limited to the curated trajectories. To generalize reasoning across large-scale sparsely annotated datasets, we formulate diagnosis as a reinforcement learning problem and optimize the model using Group-Relative Policy Optimization (GRPO)~\citep{shao2024deepseekmath}.

Given input $x = (I, q)$, the policy generates $K$ candidate responses

\begin{equation}
\{y^{(j)}\}_{j=1}^{K} = \{(CoT^{(j)}, \hat{\ell}^{(j)}, \hat{b}^{(j)})\}_{j=1}^{K},
\end{equation}

where $CoT^{(j)}$ denotes the reasoning trace, $\hat{\ell}^{(j)}$ the predicted disease label, and $\hat{b}^{(j)}$ the predicted lesion category.
Each candidate response receives reward
$
r_j = R_{\text{total}}(\mathcal{P}, y^{(j)}),
$
where $\mathcal{P}$ denotes the ground-truth diagnostic path in the taxonomy.
Rewards are normalized within groups, expressed as
$
A_j = (r_j - \mu)/\sigma.
$
The GRPO objective is defined as
{\small
\begin{equation}
\begin{aligned}
\mathcal{L}_{\text{GRPO}}(\theta) =
\mathbb{E}_{x}
\Bigg[
\frac{1}{K}\sum_{j=1}^{K}
\min(
\rho_j A_j,
\mathrm{clip}(\rho_j,1-\varepsilon,1+\varepsilon)A_j
)
\Bigg]
-
\beta
\mathrm{KL}
(
\pi_\theta(\cdot|x)
\|
\pi_{\text{ref}}(\cdot|x)
).
\end{aligned}
\end{equation}}

\paragraph{Reward Function.}

The reward function encourages diagnostically correct and structurally consistent outputs:
\begin{equation}
    R_{\text{total}} = R_{\text{format}} + R_{\text{gran}} + R_{\text{malignancy}}.
\end{equation}

\paragraph{Format Compliance Reward.}

\begin{equation}
R_{\text{format}} =
\begin{cases}
1 & \text{if required output tags are present},\\
0 & \text{otherwise}.
\end{cases}
\end{equation}

\paragraph{Granularity-wise Reward.}

\begin{equation}
R_{\text{gran}} =
\begin{cases}
0.75 w_{i^*}, & \text{if } \hat{\ell} \in \mathcal{P},\\
0, & \text{otherwise}.
\end{cases}
\label{eq:granularity}
\end{equation}
where $w_i = i/L$ represents the normalized depth of the taxonomy path. Details of $R_{\text{gran}}$ and an example are given in Appendix \cref{{app:granularity}}.

\paragraph{Malignancy Discrimination Reward.}

\begin{equation}
R_{\text{malignancy}} =
\begin{cases}
0.25, & \text{if } \hat{b} = b^\star, \\
0, & \text{otherwise},
\end{cases}
\end{equation}
where $b^\star, \hat{b} \in \{\text{benign}, \text{malignant}, \text{precancerous in situ}\}$ denote the ground-truth and predicted malignancy categories, respectively, which correspond to a fundamental clinical classification task.

In summary, \ourmethod{} integrates textbook-grounded reasoning supervision with reinforcement learning-based generalization. The combination of hierarchy-aware reasoning trajectories, supervised initialization, and GRPO-based policy refinement enables the model to produce clinically interpretable diagnostic predictions while remaining scalable to heterogeneous dermatology datasets.
\section{Experiments and Results}
\label{sec:experiments}
\paragraph{Results Summary.}
Across a diverse set of dermatology benchmarks, \ourmethod{} consistently achieves the strongest diagnostic performance among all evaluated models on both in-distribution (ID) and out-of-distribution (OOD) datasets while maintaining balanced performance on lesion condition classification. Ablation studies further show that trajectory-based SFT provides a crucial foundation for effective RL training, and that the proposed hierarchy-aware reward improves generalization to unseen disease types. Additional targeted evaluations confirm that \ourmethod{} demonstrates stronger differential diagnosis (DDx) reasoning and improved awareness of disease taxonomies compared with existing VLM baselines.

\vspace{4pt}

\noindent
\fbox{
\begin{minipage}{0.98\linewidth}
\textbf{Key Findings.}
\begin{itemize}
\item \textbf{State-of-the-art diagnostic accuracy.} \ourmethod{} achieves the highest accuracy on both ID and OOD disease diagnosis benchmarks.
\item \textbf{Balanced clinical decision performance.} The model obtains the best combined accuracy and Macro-F1 on lesion condition classification.
\item \textbf{Grounded reasoning improves RL training.} Trajectory-based SFT significantly improves structural alignment and raises the achievable performance ceiling for RL.
\item \textbf{Hierarchy-aware rewards improve generalization.} The proposed granularity-wise reward yields stronger OOD performance by encouraging hierarchical reasoning across disease categories.
\item \textbf{Improved differential diagnosis capability.} Targeted DDx and taxonomy-based evaluations confirm that \ourmethod{} more reliably distinguishes visually similar dermatological conditions.
\end{itemize}
\end{minipage}
}

\paragraph{Training \& Testing Datasets.}
We use six dermatology datasets for RL training and evaluation: PAD-UFES-20~\citep{pacheco2020impact}, DermNet~\citep{dermnet2024}, BCN20000~\citep{combalia2019bcn20000}, DERM12345~\citep{yilmaz2024derm12345}, Derm7pt~\citep{8333693}, and HAM10000~\citep{DVN_DBW86T_2018}.  
To assess cross-dataset generalization, we additionally evaluate the trained model on the dermoscopy subset of OmniMedVQA~\citep{hu2024omnimedvqa}.  
Detailed descriptions of all datasets and dataset preprocessing are provided in Appendix~\cref{appendix:dataset_details}.

\paragraph{Task Description.}
We design a visual question answering (VQA) evaluation framework to assess diagnostic reasoning under multiple distributional scenarios. The evaluation includes three settings:

\begin{enumerate}
\item \textbf{In-distribution disease diagnosis}, where the model predicts the disease category for dermoscopic images drawn from the training domain;
\item \textbf{Out-of-distribution (OOD) disease diagnosis}, which assesses generalization to unseen disease types outside the training distribution; and
\item \textbf{Lesion condition classification}, where the model classifies each case as \texttt{benign}, \texttt{malignant}, or \texttt{precancerous in situ}.
\end{enumerate}

All tasks are formulated as multiple-choice VQA problems and evaluated primarily using accuracy.  
For lesion condition classification, Macro-F1 is additionally reported to account for class imbalance.  
To ensure fair comparison, an answer extraction protocol is applied to baseline models lacking the required response format.  
Details of the multiple-choice setup, distractor design, and answer extraction protocol are provided in Appendix~\cref{appendix:dataset_details}.

\paragraph{Implementation Details.}
We adopt Qwen2.5-VL-7B-Instruct~\citep{bai2025qwen2} as the backbone model.  
Both SFT and RL stages are trained using LoRA~\citep{hu2022lora}, with hyperparameters \texttt{lora\_r=64}, \texttt{lora\_alpha=32}, and \texttt{lora\_dropout=0.1}.  
Further implementation details are provided in Appendix~\cref{appendix:implementation_details}.

\paragraph{Baselines.}
We compare \ourmethod{} with a diverse set of recent VLMs spanning general, medical, and dermatology-specific models.  
General-purpose VLMs include Qwen2.5-VL-7B and -32B~\citep{bai2025qwen2}, Qwen3-VL-8B and -32B~\citep{Qwen3-VL2025}, InternVL3-8B~\citep{zhu2025internvl3}, and LLaVA-v1.6-7B and -13B~\citep{liu2024llavanext}.  
Medical-domain VLMs include LLaVA-Med-7B~\citep{li2023llava}, HuatuoGPT-Vision-7B~\citep{chen2024huatuogpt}, and MedGemma-4B~\citep{sellergren2025medgemma}.  
We also include a dermatology-focused VLM, SkinVL-PubMM~\citep{zeng2025mm}.  

Methods such as Med-R1~\citep{lai2025med} and MedVLM-R1~\citep{pan2025medvlm} primarily rely on GRPO-style RL training.  
Therefore, the \textbf{RL without SFT} configuration in \cref{exp:ablation} can be viewed as their adaptation to dermatological diagnosis.

\subsection{\ourmethod{} Compared with SOTA Benchmarks}
\label{experiment:result_comparion}

As shown in \cref{tab:derm_vlm_results}, \textbf{\ourmethod{} outperforms all baseline models on disease diagnosis}.  
It achieves the highest average accuracy on both ID datasets (0.6385, +0.1955 over the next-best model) and OOD datasets (0.7171, +0.0284), demonstrating strong diagnostic accuracy and generalization capability.

For lesion condition classification (\cref{tab:b_or_m_accuracy_styled}), \textbf{\ourmethod{} achieves the most balanced performance}, obtaining the highest average accuracy (0.6928, +0.0297) and Macro-F1 score (0.4287, +0.0633).

\begin{table*}[t]
\centering
\caption{Comparison of \ourmethod{} with baseline LVLMs on in-distribution and OOD disease diagnostic tasks. 
\textbf{Bold} and \underline{underlined} values indicate the best and second-best results, respectively.}
\resizebox{0.98\textwidth}{!}{
\begin{tabular}{c|l|ccccccc|ccccc|c}
\toprule
\rowcolor[HTML]{E9F3FE} 
 &  & \multicolumn{7}{c|}{\textbf{In-distribution}} & \multicolumn{5}{c|}{\textbf{OOD}} &  \\ 
\cline{3-15}
\rowcolor[HTML]{E9F3FE}\multirow{-2}{*}{\textbf{Type}} & \multirow{-2}{*}{\textbf{Model}} 
& BCN20k & HAM10k & PAD & Derm12345 & Derm7pt & DermNet & Avg. 
& ISBI16 & ISIC18 & ISIC19 & ISIC20 & Monk22 & Avg. \\
\midrule \midrule
\multirow{3}{*}{\textbf{General}} 
& Qwen2.5-VL-7B-Instruct & 0.4911 & \underline{0.5162} & 0.5033 & 0.3521 & 0.2962 & 0.4852 & 0.4367 & 
0.3655 & 0.6301 & 0.6359 & 0.3325 & \textbf{0.7078}  & 0.5027 \\
& Qwen3-VL-8B-Instruct & 0.2947 & 0.2085 & 0.3275 & 0.2386 & 0.2354 & 0.2899 & 0.2585 
& \underline{0.6996} & 0.2890 & 0.3934 & \textbf{0.8837} & 0.4610  & 0.5939 \\
& InternVL3-8B & \underline{0.4996} & 0.4705 & \underline{0.6009} & 0.3453 & 0.3364 & \underline{0.5473}  & \underline{0.4430}
& 0.4580 & \underline{0.6474} & \textbf{0.7746} & 0.3531 & \underline{0.6948}  & 0.5841 \\
& LLaVA-v1.6-7B & 0.2299 & 0.2416 & 0.3080 & 0.3046 & 0.2354 & 0.2885 & 0.2636 
& 0.5819 & 0.1908 & 0.3134 & 0.5899 & 0.3312  & 0.4351 \\
& LLaVA-v1.6-13B & 0.3933 & 0.3779 & 0.4469 & \underline{0.4121} & 0.3376 & 0.4193  & 0.3960 
& 0.2563 & 0.1676 & 0.3415 & 0.1584 & 0.2143  & 0.2560\\
& Qwen2.5-VL-32B-Instruct & 0.4746 & 0.4057 & 0.4469 & 0.3815 & \underline{0.3596} & \textbf{0.5763}  & 0.4307
& 0.5966 & 0.2601 & 0.3445 & 0.6040 & 0.3247  & 0.4587 \\
& Qwen3-VL-32B-Instruct & 0.3156 & 0.2363 & 0.3688 & 0.2801 & 0.2536 & 0.3304  & 0.2892 
& \textbf{0.6996} & 0.3353 & 0.5596 & \underline{0.8408} & 0.6688  & 0.6651 \\
\midrule
\multirow{3}{*}{\textbf{Medical}} 
& LLaVA-Med-7b & 0.2508 & 0.2091 & 0.2039 & 0.2503 & 0.2135 & 0.2870  & 0.2400  
& 0.2542 & 0.2601 & 0.3928 & 0.5314 & 0.4545  & 0.4170 \\
& HuatuoGPT-Vision-7B & 0.4799 & 0.4944 & 0.4295 & 0.3638 & 0.2561 & 0.4719 & 0.4241 & 0.4202 & 0.5896 & 0.5657 & 0.4505 & 0.5779  & 0.5101 \\
& MedGemma-4B & 0.4130 & 0.2627 & 0.3753 & 0.3767 & 0.2924 & 0.3240 & 0.3548 & 0.5168 & 0.5318 & 0.7214 & 0.7508 & 0.5584  & \underline{0.6887} \\
\midrule
\multirow{2}{*}{\textbf{Dermatology}} 
&SkinVL-PubMM & - & - & - & - & - &-  & - & - & - & - &- & -  & -\\
& \cellcolor[HTML]{DAE0FB}\textbf{\ourmethod} & \cellcolor[HTML]{DAE0FB}\textbf{0.6345} & \cellcolor[HTML]{DAE0FB}\textbf{0.7214} & \cellcolor[HTML]{DAE0FB}\textbf{0.6573} & \cellcolor[HTML]{DAE0FB}\textbf{0.6608} & \cellcolor[HTML]{DAE0FB}\textbf{0.4955} & \cellcolor[HTML]{DAE0FB}0.5370 & \cellcolor[HTML]{DAE0FB}\textbf{0.6385} 
& \cellcolor[HTML]{DAE0FB}0.6828 & \cellcolor[HTML]{DAE0FB}\textbf{0.7630} & \cellcolor[HTML]{DAE0FB}\underline{0.7630} & \cellcolor[HTML]{DAE0FB}0.6708 & \cellcolor[HTML]{DAE0FB}0.6494 & \cellcolor[HTML]{DAE0FB}\textbf{0.7171}  \\
\bottomrule
\end{tabular}}
\label{tab:derm_vlm_results}
\end{table*}

\begin{table*}[htbp]
\centering
\scriptsize
\caption{Comparison of \ourmethod{} with baseline LVLMs on in-distribution skin lesion diagnostic tasks. 
 \textbf{Bold} and \underline{underlined} values indicate the best and second-best results, respectively.}
\label{tab:b_or_m_accuracy_styled}
\resizebox{0.98\textwidth}{!}{
\begin{tabular}{c|l|cccccccccccc|cc}
\toprule
\rowcolor[HTML]{E9F3FE}
 &  & \multicolumn{14}{c}{\textbf{In-distribution}} \\ 
\cline{3-16}
\rowcolor[HTML]{E9F3FE}
\multirow{-2}{*}{\textbf{Type}} & \multirow{-2}{*}{\textbf{Model}} 
& \multicolumn{2}{c}{BCN20k} & \multicolumn{2}{c}{HAM10k} & \multicolumn{2}{c}{PAD} 
& \multicolumn{2}{c}{Derm12345} & \multicolumn{2}{c}{Derm7pt} & \multicolumn{2}{c}{DermNet} 
& \multicolumn{2}{c}{Avg.} \\ 
\cline{3-16}
\rowcolor[HTML]{E9F3FE}
 &  & Acc & F1 & Acc & F1 & Acc & F1 & Acc & F1 & Acc & F1 & Acc & F1 & Acc & F1 \\
\midrule
\multirow{3}{*}{\textbf{General}} 
& Qwen2.5-VL-7B-Instruct & 0.4308 & 0.3346 & 0.2905 & 0.2449 & 0.4100 & 0.3632 & 0.3046 & 0.2045 & 0.3661 & 0.3468 & 0.5399 & 0.4390  & 0.3698 & 0.2910 \\
& Qwen3-VL-8B-Instruct & 0.4581 & 0.3105 & 0.6592 & 0.3601 & 0.3124 & 0.2891 & 0.7437 & 0.3716 & 0.6494 & \underline{0.5858} & 0.5059 & 0.3503  & 0.5924 & 0.3649  \\
&InternVL3-8B& 0.4795 & \textbf{0.3768} & 0.3514 & 0.2639 & \underline{0.5033} & \textbf{0.4932} & 0.4769 & 0.3091 & 0.4334 & 0.4343 & \underline{0.5932} & \textbf{0.5366}  & 0.4619 & 0.3610\\
& LLaVA-v1.6-7B & 0.4014 & 0.3036 & 0.5897 & \underline{0.3669} & 0.2907 & 0.2743 & 0.6414 & 0.3285 & 0.5356 & 0.5361 & 0.4601 & 0.3109  & 0.5174 & 0.3428  \\
& LLaVA-v1.6-13B & 0.3704 & 0.3070 & 0.4057 & 0.2888 & 0.3536 & 0.3458 & 0.3847 & 0.2330 & 0.3739 & 0.4181 & 0.4267 & 0.3524  & 0.3849 & 0.2978\\
& Qwen2.5-VL-32B-Instruct & \underline{0.4799} & 0.3507 & 0.4917 & 0.3390 & 0.3514 & 0.3027 & 0.6093 & 0.3442 & 0.4903 & 0.4948 & 0.5896 & 0.4511  & 0.5231 & \underline{0.3654}\\
& Qwen3-VL-32B-Instruct & 0.4481 & 0.2452 & \underline{0.7598} & 0.3508 & 0.2560 & 0.1968 & 0.8716 & \underline{0.3923} & 0.6986 & 0.4991 & 0.5333 & 0.3092  & 0.6491 & 0.3337 \\
\midrule
\multirow{3}{*}{\textbf{Medical}} 
& LLaVA-Med-7b & 0.4465 & 0.2058 & \textbf{0.7968} & 0.2956 & 0.2299 & 0.1246 & \textbf{0.8994} & 0.3157 & \underline{0.6986} & 0.4113 & 0.5459 & 0.2354  & \underline{0.6631} & 0.2714 \\
& HuatuoGPT-Vision-7B & 0.4489 & \underline{0.3525} & 0.3349 & 0.2858 & \textbf{0.5098} & \underline{0.4725} & 0.3682 & 0.2390 & 0.3842 & 0.3957 & 0.5562 & \underline{0.4680}  & 0.4105 & 0.3267 \\
& MedGemma-4B & 0.0817 & 0.0746 & 0.0668 & 0.0558 & 0.3384 & 0.2302 & 0.0692 & 0.0574 & 0.0517 & 0.0813 & 0.1746 & 0.1531  & 0.0942 & 0.0816 \\
\midrule
\multirow{2}{*}{\textbf{Dermatology}} 
& SkinVL-PubMM  & 0.4368 & 0.2373 & 0.7187 & 0.3459 & 0.2560 & 0.1733 & 0.8740 & 0.3196 & 0.6779 & 0.5491 & 0.5592 & 0.2816  & 0.6392 & 0.3100 \\
 & \cellcolor[HTML]{DAE0FB}\textbf{\ourmethod} & \cellcolor[HTML]{DAE0FB}\textbf{0.5370} 
& \cellcolor[HTML]{DAE0FB}0.3486
& \cellcolor[HTML]{DAE0FB}0.7472
& \cellcolor[HTML]{DAE0FB}\textbf{0.4039}
& \cellcolor[HTML]{DAE0FB}0.3536 
& \cellcolor[HTML]{DAE0FB}0.2888
& \cellcolor[HTML]{DAE0FB}\underline{0.8918}
& \cellcolor[HTML]{DAE0FB}\textbf{0.4786}
& \cellcolor[HTML]{DAE0FB}\textbf{0.7089}
& \cellcolor[HTML]{DAE0FB}\textbf{0.6589}
& \cellcolor[HTML]{DAE0FB}\textbf{0.6257}
& \cellcolor[HTML]{DAE0FB}0.4271
& \cellcolor[HTML]{DAE0FB}\textbf{0.6928}
& \cellcolor[HTML]{DAE0FB}\textbf{0.4287} \\

\bottomrule
\end{tabular}}
\end{table*}

Several baseline models exhibit notable biases.  
MedGemma-4B shows a strong bias toward the rare class \texttt{C: precancerous in situ} (7,860 of 8,390 predictions), leading to poor overall performance.  
LLaVA-Med-7B demonstrates a bias toward the majority class \texttt{A: benign} (5,563 of 8,390 predictions), resulting in low F1 scores.  
SkinVL-PubMM also shows a mild bias toward the \texttt{A: benign} category.  

Moreover, SkinVL-PubMM rarely produces meaningful diagnostic responses on both ID and OOD tasks even when prompts are simplified. Consequently, its metrics are reported as “$-$”, indicating that the model fails to produce meaningful answers rather than being affected by the answer extraction protocol.

\subsection{Ablation Study}
\label{exp:ablation}

\begin{figure*}[htbp]
\centering
\includegraphics[width=1\textwidth]{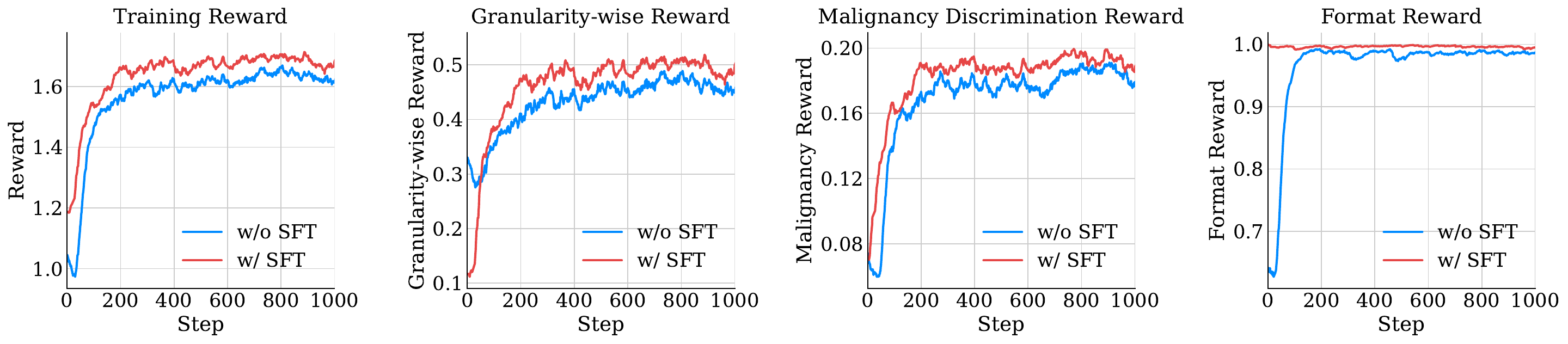}
\caption{Training reward curve comparison of \ourmethod{} with or without SFT stage.}
\label{fig:train_reward_curve}
\end{figure*}

We summarize ablation results evaluating trajectory SFT, RL stages, and reward design in \cref{tab:diagnosis_accuracy_ablation_final} and~\cref{tab:b_or_m_accuracy_ablation_final}.
\begin{enumerate}
\item
\textbf{Stage 1 (Trajectory SFT) Improves Diagnostic Ability and Structural Alignment.}
The \textbf{SFT} model outperforms the \textbf{base model} Qwen2.5-VL-7B-Instruct overall.  
The base model also frequently fails to follow the required output format. Under strict answer extraction, it fails to produce valid answers in approximately 30\% of ID diagnosis cases, compared with only 2\% for the SFT model.  
These results highlight the critical role of trajectory-based SFT in establishing grounded diagnostic reasoning and structural alignment.
\item
\textbf{Stage 2 (RL) Significantly Boosts Performance and Generalization.}
The full \ourmethod{} pipeline achieves markedly higher performance than earlier checkpoints and the base model.  
The proposed reward design encourages structured hierarchical reasoning and lesion differentiation.  
Strong OOD performance further confirms improved generalization.  
We select the checkpoint at 1,500 RL steps as the representative model.
\item
\textbf{High-Quality SFT is Essential for Effective RL.}
The complete \ourmethod{} pipeline consistently outperforms the \textbf{RL without SFT} variant.  
Although RL without SFT improves upon the base model, its best performance remains significantly lower (\cref{fig:train_reward_curve}).  
This indicates that RL-only approaches lack the grounded knowledge necessary for robust reasoning.
\item
\textbf{Granularity-wise Reward Improves Generalization Ability.}
Replacing the granularity-aware reward with a standard binary reward leads to weaker OOD performance.  
This suggests that the proposed reward provides more informative learning signals by encouraging hierarchical reasoning across the disease taxonomy.
\end{enumerate}

\begin{table*}[htbp!]
\centering
\caption{Ablation study of \ourmethod{} across in-distribution and OOD datasets. 
\textbf{Bold} and \underline{underlined} indicate the best and second-best performance, respectively.}
\resizebox{0.98\textwidth}{!}{
\begin{tabular}{l|ccccccc|ccccc|c}
\toprule
\rowcolor[HTML]{E9F3FE}
 & \multicolumn{7}{c|}{\textbf{In-distribution}} & \multicolumn{5}{c|}{\textbf{Out-of-distribution (OOD)}} &  \\ 
\cline{2-14}
\rowcolor[HTML]{E9F3FE}\multirow{-2}{*}{\textbf{Model}} 
& BCN20k & HAM10k & PAD & Derm12345 & Derm7pt & DermNet & Avg. 
& ISBI16 & ISIC18 & ISIC19 & ISIC20 & Monk22 & Avg. \\
\midrule \midrule
Qwen2.5-VL-7B-Instruct 
& 0.4911 & 0.5162 & 0.5033 & 0.3521 & 0.2962 & 0.4852 & 0.4367 
& 0.3655 & 0.6301 & 0.6359 & 0.3325 & \underline{0.7078}  & 0.5027 \\
SFT
& 0.4899 & 0.6122 & 0.5358 & 0.4181 & 0.2937 & 0.4793  & 0.4743
& 0.5693 & 0.4509 & 0.5651 & 0.4249 & 0.6039  & 0.5153 \\
RL without SFT & 0.5898 & 0.6671 & 0.5813 & 0.6056 & 0.3855 & 0.5296  & 0.5843 & 0.4223 & \underline{0.7514} & 0.7208 & 0.3663 & \textbf{0.7597}  & 0.5674 \\
RL with standard reward & \underline{0.6252} & \textbf{0.7227} & 0.6182 & \textbf{0.6628} & \textbf{0.5265} & \textbf{0.5444}  & \underline{0.6379}  
& \textbf{0.6870} & 0.7399 & 0.6885 & 0.5528 & 0.5195  & 0.6386 \\
\ourmethod{} (500 steps) 
& 0.6139 & 0.6870 & 0.6095 & 0.6020 & 0.3351 & 0.4882  & 0.5875
& 0.6744 & 0.6474 & 0.6805 & \textbf{0.6848} & 0.5519  & 0.6742 \\
\ourmethod{} (1,000 steps) 
& 0.6083 & 0.7009 & \underline{0.6508} & 0.6487 & 0.4502 & 0.4956  & 0.6156 
& 0.6660 & 0.7168 & \textbf{0.7648} & 0.5982 & 0.5909  & \underline{0.6870} \\
\ourmethod{} (1,500 steps) 
& \textbf{0.6345} & \underline{0.7214} & \textbf{0.6573} & \underline{0.6608} & \underline{0.4955} & \underline{0.5370}  & \textbf{0.6385} 
& \underline{0.6828} & \textbf{0.7630} & \underline{0.7630} & \underline{0.6708} & 0.6494  & \textbf{0.7171}  \\ 
\bottomrule
\end{tabular}}
\label{tab:diagnosis_accuracy_ablation_final}
\end{table*}

\begin{table*}[htbp]
\centering
\caption{Ablation study of \ourmethod{} on skin lesion diagnostic task across in-distribution datasets. \textbf{Bold} and \underline{underlined} indicate the best and second-best performance, respectively.}
\label{tab:b_or_m_accuracy_ablation_final}
\resizebox{0.98\textwidth}{!}{
\begin{tabular}{l|cccccccccccc|cc}
\toprule
\rowcolor[HTML]{E9F3FE}
 & \multicolumn{14}{c}{\textbf{In-distribution}} \\ 
\cline{2-15}
\rowcolor[HTML]{E9F3FE}
\multirow{-2}{*}{\textbf{Model}} 
& \multicolumn{2}{c}{BCN20k} & \multicolumn{2}{c}{HAM10k} & \multicolumn{2}{c}{PAD} 
& \multicolumn{2}{c}{Derm12345} & \multicolumn{2}{c}{Derm7pt} & \multicolumn{2}{c}{DermNet} 
& \multicolumn{2}{c}{Avg.} \\ 
\rowcolor[HTML]{E9F3FE}
 & Acc & F1 & Acc & F1 & Acc & F1 & Acc & F1 & Acc & F1 & Acc & F1 & Acc & F1 \\
\midrule
Qwen2.5-VL-7B-Instruct 
& 0.4308 & 0.3346 & 0.2905 & 0.2449 & \underline{0.4100} & \underline{0.3632} & 0.3046 & 0.2045 & 0.3661 & 0.3468 & 0.5399 & 0.4390  & 0.3698 & 0.2910 \\
SFT 
& 0.4972 & 0.3228 & 0.6373 & 0.3567 & 0.3688 & 0.2970 & 0.8072 & 0.4037 & 0.6223 & 0.5876 & 0.6006 & 0.4590  & 0.6271 & 0.3868 \\
RL without SFT & 0.4819 & \textbf{0.3654} & 0.3812 & 0.3037 & \textbf{0.4338} & \textbf{0.3740} & 0.4748 & 0.2839 & 0.3946 & 0.3824 & 0.5962 & \underline{0.4792}  & 0.4602 & 0.3414 \\
RL with standard reward 
& \textbf{0.5407} & \underline{0.3633} & 0.6810 & 0.3849 & 0.3991 & 0.3232 & 0.8378 & 0.4480 & 0.6624 & 0.6416 & \textbf{0.6450} & \textbf{0.4841}  & 0.6658 & \underline{0.4254} \\
\ourmethod (500 steps)
 & 0.5229 & 0.3305 & \textbf{0.7565} & 0.3867 & 0.3362 & 0.2638 & \textbf{0.8994} & \underline{0.4584} & \textbf{0.7413} & \underline{0.6571} & 0.6036 & 0.4215  & \underline{0.6928} & 0.4123 \\
\ourmethod (1,000 steps)
& 0.5342 & 0.3528 & 0.6962 & \underline{0.3880} & 0.3601 & 0.2916 & 0.8519 & 0.4421 & 0.6792 & 0.6386 & 0.6228 & 0.4325  & 0.6684 & 0.4150 \\
\ourmethod (1,500 steps)
& \underline{0.5370} & 0.3486 & \underline{0.7472} & \textbf{0.4039} & 0.3536 & 0.2888 & \underline{0.8918} & \textbf{0.4786} & \underline{0.7089} & \textbf{0.6589} & \underline{0.6257} & 0.4271  & \textbf{0.6928} & \textbf{0.4287} \\
\bottomrule
\end{tabular}}
\end{table*}

\subsection{Targeted DDx and Hierarchical Evaluation}
\label{experiment:ddx_hierarchical}
\begin{wrapfigure}{r}{0.5\textwidth}
\vspace{-30pt}
\centering
\includegraphics[width=\linewidth]{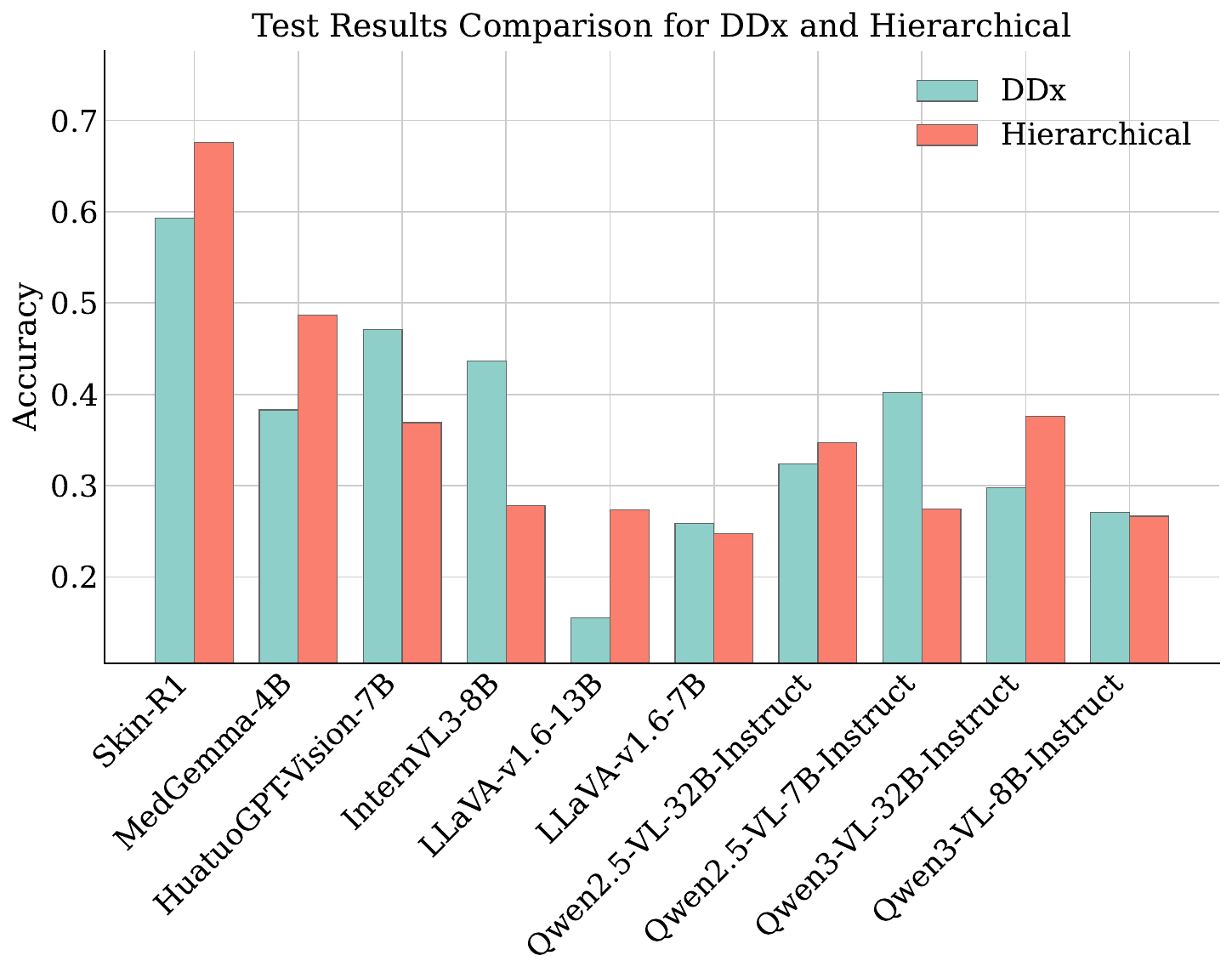}
\caption{Comparison on DDx and hierarchical diagnosis tasks.}
\label{fig:ddx_hierarchical}
\vspace{-40pt}
\end{wrapfigure}
To evaluate DDx and hierarchy-aware reasoning, we construct two targeted evaluation tasks: (1) \emph{DDx diagnosis} and (2) \emph{hierarchical diagnosis}.  
These tasks modify the standard ID benchmark by replacing distractors using the disease taxonomy $\mathcal{T}_{tax}$ and the differential diagnosis graph $\mathcal{G}_{DDx}$.

Evaluation uses held-out images never seen during training. The taxonomy and DDx graph are used only to generate distractors and are not provided to the model during inference.

\begin{itemize}
\item For hierarchical diagnosis, distractors correspond to ancestor nodes of the ground-truth diagnosis.
\item For DDx diagnosis, distractors correspond to confusable neighboring diseases in $\mathcal{G}_{DDx}$.
\end{itemize}

As shown in \cref{fig:ddx_hierarchical}, \ourmethod{} achieves the highest accuracy in both settings, demonstrating stronger differentiation of visually similar diseases and improved awareness of taxonomic relationships.

\subsection{Qualitative Analysis}
\label{sec:qualitative}

Qualitative comparisons (Appendix~\cref{appendix:qualitative_analysis}) show that \ourmethod{} produces responses that are more concise, accurate, and consistent with expert diagnostic reasoning compared with the base Qwen2.5-VL-7B model.

For example, in Open-ended Case~1 of Appendix~\cref{appendix:qualitative_analysis}, \ourmethod{} correctly concludes that the observed features ``indicate a benign pigmented nevus rather than melanoma'', demonstrating precise DDx reasoning.  
In contrast, the base model fails to capture this critical clinical distinction.

While the base model tends to produce longer descriptive outputs, it often sacrifices diagnostic precision.  
\ourmethod{}, in contrast, produces more focused and reliable diagnoses, reflecting stronger grounded diagnostic reasoning capabilities.

\section{Summary and Discussion}
\label{sec:conclusion}

\paragraph{\textbf{Summary.}} We addressed three major challenges that limit the trustworthiness
and clinical utility of dermatological vision–language models: dataset heterogeneity, the scarcity of grounded reasoning supervision, and the difficulty of transferring complex diagnostic reasoning to large-scale datasets with sparse annotations.

We introduced \ourmethod{}, a dermatology-oriented VLM trained through a three-stage framework that integrates textbook-grounded supervision with reinforcement learning. First, we construct a textbook-based reasoning generator that synthesizes hierarchy-aware and differential-diagnosis (DDx)-informed diagnostic trajectories, providing expert-aligned supervision. Second, these trajectories are used for supervised fine-tuning (SFT), establishing a grounded reasoning foundation. Third, we introduce a reinforcement learning framework that leverages hierarchical diagnostic rewards to generalize these reasoning patterns to large-scale sparsely annotated datasets.

Extensive experiments across multiple dermatology benchmarks show that \ourmethod{} consistently outperforms state-of-the-art baselines in diagnostic accuracy and robustness. Ablation studies further highlight the critical role of trajectory-based SFT in enabling effective reinforcement learning and improving generalization.

Overall, our results highlight the importance of grounded reasoning supervision for developing trustworthy medical VLMs and suggest a promising direction toward clinically reliable multimodal diagnostic systems.

\paragraph{\textbf{Discussion.}}
Although Skin-R1 improves diagnostic reasoning and fine-grained prediction, several limitations remain before reliable real-world deployment.

First, the reward signal and evaluation is answer-level: models are scored by multiple-choice accuracy, while the correctness and faithfulness of the generated reasoning traces are examined only qualitatively; quantifying rationale quality in more open-ended, clinically realistic settings, together with confidence calibration, abstention or referral under high uncertainty, and severity-aware error analysis for malignant conditions, is left to future work.

Second, our pipeline extracts a taxonomy and DDx graph offline and keeps them fixed during training; since these structures also generate the distractors for the in-distribution and targeted evaluations, those results partially reflect consistency with this fixed clinical knowledge, whereas the out-of-distribution evaluation (OmniMedVQA), whose options are independent of our taxonomy, is not subject to this coupling. This static design also limits adaptation to evolving or dataset-specific ontologies.

Third, Skin-R1 is instantiated only on Qwen2.5-VL-7B; whether the same SFT+RL recipe transfers to other backbones, including medical ones, and how it compares with additional dermatology-specific foundation models, remains to be studied.

Finally, we observe modest degradation in certain Fitzpatrick skin types (Appendix~\cref{appendix:Fitzpatrick}), likely reflecting training-data bias, and our outcome-level objective does not explicitly reward intermediate reasoning steps; fairness-aware training, domain adaptation, and process-level supervision are promising directions; however, defining reliable reward signals for clinical reasoning remains challenging.

\section{Acknowledgements.}  The work of Vasant G Honavar and Zehao Liu was supported in part by grants from the National Science Foundation (2226025, 2225824), the National Center for Advancing
Translational Sciences, and the National Institutes of Health (UL1 TR002014).


 
\newpage
%
%
\bibliographystyle{splncs04}
\bibliography{main}
\newpage
\appendix

\appendix
\renewcommand{\theHsection}{A\arabic{section}}

\section*{Appendix}
\section{Details of Datasets}
\label{appendix:dataset_details}

We use six datasets for reinforcement learning (RL) training: PAD-UFES-20~\citep{pacheco2020impact}, DermNet~\citep{dermnet2024}, BCN20000~\citep{combalia2019bcn20000}, DERM12345~\citep{yilmaz2024derm12345}, Derm7pt~\citep{8333693}, and HAM10000~\citep{DVN_DBW86T_2018}.

\vspace*{10pt}
\noindent\textbf{PAD-UFES-20}~\citep{pacheco2020impact}
PAD-UFES-20 is a dermoscopic and clinical skin image dataset collected by the Federal University of Espírito Santo (UFES), Brazil. It contains 2,298 images across six lesion types captured by mobile devices in real clinical settings. In addition to images, it provides 22 clinical and demographic attributes (e.g., age, sex, anatomical site).

\vspace*{10pt}
\noindent\textbf{DermNet}~\citep{dermnet2024}
DermNet is a large-scale clinical dermatology image collection with approximately 19,500 images spanning 23 coarse and more than 600 fine-grained disease categories. Each image is annotated with rich meta-features (23 total). To reduce noise and improve training efficiency, we preprocess DermNet by filtering out approximately 80\% of samples: we remove many rare long-tail diseases and retain categories relevant to skin cancers and their differential diagnoses, consistent with the scope of this study.

\vspace*{10pt}
\noindent\textbf{BCN20000}~\citep{combalia2019bcn20000}
BCN20000 was introduced by Hospital Cl\'inic de Barcelona as a curated dermoscopic dataset containing 18,946 images from 8 diagnostic classes. It includes clinical metadata such as patient age, sex, and lesion location, and serves as a high-quality benchmark for automated lesion classification and melanoma detection.

\vspace*{10pt}
\noindent\textbf{DERM12345}~\citep{yilmaz2024derm12345}
DERM12345 is a hierarchical dermatology dataset consisting of 12,345 dermoscopic and clinical images. It provides multi-granular categorization across three taxonomy levels (5/15/40 classes), enabling evaluation of hierarchical reasoning beyond flat labels.

\vspace*{10pt}
\noindent\textbf{Derm7pt}~\citep{8333693}
Derm7pt is a publicly available dermoscopic dataset containing 1,011 images annotated according to the 7-point melanoma checklist. It includes seven clinical feature annotations (e.g., atypical network, blue-white veil, streaks, regression structures) and lesion labels for 20 diagnostic categories.

\vspace*{10pt}
\noindent\textbf{HAM10000}~\citep{DVN_DBW86T_2018}
HAM10000 (``Human Against Machine with 10,000 Training Images'') is a widely used dermoscopic dataset containing 10,015 images from 7 diagnostic categories collected from multiple sources. It includes both benign and malignant lesions, such as melanoma, melanocytic nevus, basal cell carcinoma, actinic keratosis, benign keratosis, dermatofibroma, and vascular lesions.

Each dataset is divided into training/validation/test subsets using either official splits or, when unavailable, a random 6:2:2 split. The validation set is not used for training or evaluation; it is retained only for compatibility with datasets that provide a three-way split. In total, the training set contains 26,507 samples and the test set contains 8,390 samples.

\paragraph{Multiple-choice construction for in-distribution datasets.}
To construct multiple-choice options consistent with hierarchical taxonomy, each instance draws candidates from both the dataset-specific taxonomy and the shared global taxonomy. Options are formed by mixing (i) correct on-path diagnoses sampled at different depths and (ii) incorrect distractors sampled from diseases not on the ground-truth path, yielding four candidate options per question. For targeted DDx and hierarchical tests, we collect all available DDx neighbors and taxonomy-path diagnoses as candidates, resulting in a variable number of options. Cases that cannot form at least two valid options are excluded.

\paragraph{Evaluation protocol for in-distribution datasets.}
For disease diagnosis, the correct answer is defined as the most fine-grained candidate that lies on the ground-truth taxonomy path. For lesion condition diagnosis, we evaluate the same test instances but replace the options with \texttt{A: benign}, \texttt{B: malignant}, and \texttt{C: precancerous in situ}. Sample prompt templates along with \ourmethod{} generated responses are provided in Appendix~\cref{appendix:prompts}.

\paragraph{Implementation of the granularity-wise reward.}
During RL training, the granularity-wise reward $R_{\text{gran}}$ follows the procedure in the \textbf{Granularity-wise Reward} subsection of \cref{sec:method}. Specifically, if a predicted option lies on the correct taxonomy path, its reward is determined by its depth in the global taxonomy; otherwise it is set to 0.

\paragraph{Answer extraction.}
To ensure fair evaluation, we apply a ten-level answer extraction protocol to baseline and SOTA model outputs. The protocol progressively relaxes matching criteria, ranging from strict format parsing (e.g., exact tag-based extraction) to semi-structured matching (option-letter and option-text matching), and finally to permissive keyword-based extraction. This hierarchy reduces differences caused by format compliance rather than prediction quality. We additionally perform manual spot checks to verify that the protocol captures all meaningful answers.
We perform a quantitative analysis on the extraction validity.
As shown in \cref{tab:model_validity}, all models achieve $>$99\% valid extraction except Qwen3-VL-8B (6.31\% invalid due to output truncation), LLaVA-Med and SkinVL-PubMM (2.1\% and 42.88\% invalid due to meaningless answer) .
\begin{table}[htbp]
\caption{Answer extraction validity} 
  \label{tab:model_validity}
  \centering
  \begin{tabular}{lc}
  \hline
  Model & Extraction Validity (\%) \\
  \hline
  \ourmethod{} & 100.00 \\
  Qwen2.5-VL-7B & 100.00 \\
  InternVL3-8B & 99.99 \\
  Qwen3-VL-32B & 99.98 \\
  HuatuoGPT-Vision-7B & 99.94 \\
  MedGemma-4B & 99.88 \\
  LLaVA-v1.6-7B & 99.82 \\
  Qwen2.5-VL-32B & 99.82 \\
  LLaVA-v1.6-13B & 99.64 \\
  LLaVA-Med-v1.5 & 97.90 \\
  Qwen3-VL-8B & 93.69 \\
  SkinVL-PubMM & 57.12 \\
  \hline
  \end{tabular}
\end{table}

\paragraph{Out-of-distribution (OOD) disease diagnosis dataset.}
In addition to the test subsets above, we evaluate generalization using the OmniMedVQA dermoscopy subset~\citep{hu2024omnimedvqa}. OmniMedVQA is a large-scale medical VQA benchmark assembled from 73 public datasets across 12 imaging modalities (e.g., dermoscopy, MRI, CT, X-ray, histopathology), containing 118,010 images and 127,995 QA items converted to multi-choice format.

This dermoscopy subset aggregates multiple public datasets, including Fitzpatrick17k~\citep{groh2021evaluating}, ISBI2016~\citep{codella2018skin}, ISIC2018~\citep{codella2019skin}, ISIC2019~\citep{ISIC2019Challenge}, ISIC2020~\citep{rotemberg2021patient}, Monkeypox Skin Image 2022~\citep{islam2022web}, and PAD-UFES-20~\citep{pacheco2020impact}. To evaluate diagnostic capability, we use only samples with \texttt{"question\_type": "Disease Diagnosis"} and exclude \texttt{"Modality Recognition"} and \texttt{"Anatomy Identification"}. The Fitzpatrick17k subset is used for skin tone fairness evaluation. PAD-UFES-20 is excluded to avoid data duplication.

\section{Details of Textbook Knowledge Extraction}
\label{appendix:data_extraction}

\subsection{Statistical Information}
We collect 220 diagnostic exemplars $(I_i, r_i, d_i)$, each consisting of a clinical image $I_i$, a textual rationale $r_i$ describing diagnostic evidence and reasoning, and a confirmed diagnosis $d_i$.

The differential diagnosis (DDx) graph $\mathcal{G}_{\text{DDx}}$ contains 211 nodes and 245 edges; 61.36\% of exemplars are covered (i.e., have at least one DDx neighbor). The disease taxonomy $\mathcal{T}_{\text{tax}}$ contains 458 nodes and 473 edges. In addition, we curate 192 synonym mappings to align labels across resources.

The DDx graph and taxonomy jointly cover all diagnostic labels in the RL datasets (in-distribution setting). Specifically, original RL labels are first expanded to full diagnostic names and then mapped to corresponding nodes in $\mathcal{G}_{\text{DDx}}$ and $\mathcal{T}_{\text{tax}}$.

\subsection{Textbook Knowledge Extraction Pipeline}
We use Part 20 (Neoplasia) of \emph{Fitzpatrick's Dermatology, 9e}~\citep{kang2019fitzpatrick} (399 pages) as the source, covering neoplasia and related DDx diseases aligned with our study scope and the RL training datasets.

Our extraction pipeline is adapted from~\citep{kim2024transparent} and is largely automatic, with minimal human intervention restricted to selecting relevant image clusters.

\paragraph{Image and Text Block Extraction}
We first extract images and page text blocks from the PDF preserving metadata including the page number and spatial bounding box coordinates of each element to retain their original layout positions. To retain lesion-focused figures while filtering non-lesion content (e.g., diagrams and charts), we subsequently apply unsupervised clustering and filtering (below).

\paragraph{Image Clustering and Filtering}
To reduce non-lesion content (e.g., diagrams, flowcharts, histology), we perform unsupervised image clustering and filter by cluster labels. We compute visual embeddings using a pretrained EfficientNet backbone, then apply two-level hierarchical k-means clustering (4 clusters per level). Clusters corresponding to non-clinical content are excluded via cluster-level inspection.

\paragraph{Text Pairing and Filtering}
We perform rule-based image--text alignment to identify captions and contextual references. Candidate text blocks on the same page are ranked using geometric proximity features from bounding box relations. Caption identification combines spatial proximity with regular expression matching of figure identifiers (e.g., ``Figure'', ``Fig.''). We also retrieve cross-page contextual mentions of the same figure identifier within a $\pm 2$ page window. Each image is paired with a structured tuple: figure label, caption text, and contextual references. In parallel, DDx and taxonomy signals are extracted by detecting DDx-related phrases (e.g., ``differential diagnosis'', ``DDx'') and by parsing document structure (chapter/section titles) for hierarchical organization.

\paragraph{Diagnostic Rule Extraction}
For each image--text tuple, we use an LLM (gpt-4.1-mini) to extract diagnostic rules and diagnosis outcomes, producing rule-like statements that map visual findings to diagnostic conclusions without introducing external knowledge. Outputs are parsed into $\mathcal{D}=\{(I_i, r_i, d_i)\}_{i=1}^N$, and entries lacking valid rule structures are discarded.

\paragraph{DDx and Taxonomy Confirming and Refining}
We refine DDx and taxonomy representations via LLM-assisted normalization and validation. For DDx construction, each subject disease is paired with a curated list of clinically grounded differentials (excluding pathology-based differentials), yielding a directed graph where nodes are diseases and edges encode subject-to-differential relations. For taxonomy construction, entries are mapped to book pages and associated chapter titles to form hierarchical paths (e.g., Chapter $\rightarrow$ Diagnosis). The resulting $\mathcal{G}_{\text{DDx}}$ and $\mathcal{T}_{\text{tax}}$ are stored in structured formats, forming an integrated resource of image-grounded captions, diagnostic rules, DDx structure, and hierarchical disease organization.

\begin{promptbox}{Prompt template for diagnostic rule extraction}
You are a dermatology expert. The following text may or may not describe a diagnostic rule. \\\\
Your task:
\\
- Extract and rewrite only the diagnostic rule(s) for the condition described in the text, based on symptoms or visual signs.
\\
- Focus on how a physician would reason from visual symptoms to diagnosis, using specific features or patterns.
\\
- Rewrite the rule(s) in a concise, declarative format (not explanatory or descriptive).
\\
- Do not include any information that is not contained in the original text.
\\
- Avoid vague phrases such as "aids diagnosis", "these features are characteristic", or general comments about appearance.
\\\\
Example:

<Diagnosis rule>Presence of a rough surface and "stuck-on" appearance indicates seborrheic keratosis.</Diagnosis rule> <Diagnosis result>seborrheic keratosis</Diagnosis result>
\\\\
Formatting rules:
\\
- If there is a valid diagnostic rule: return it in  
'<Diagnosis rule>...</Diagnosis rule>' and the corresponding diagnosis in '<Diagnosis result>...</Diagnosis result>'
\\
- If no diagnostic rule is present, return: '<Diagnosis rule> NOT A RULE </Diagnosis rule>' '<Diagnosis result></Diagnosis result>'
\\\\
Original text:\{text\}
\\\\
Output format:<Diagnosis rule>...</Diagnosis rule><Diagnosis result>...</Diagnosis result>
\end{promptbox}

\begin{promptbox}{Prompt template for DDx extraction}
You are a dermatology expert. The following text is extracted from a dermatology textbook and contains information about differential diagnosis (DDx).
\\\\
Your task:
\\
- Identify and extract the differential diagnosis (DDx) list for a clearly defined subject disease.
\\
- Only include DDx based on symptoms or visual/clinical signs.
\\
- Do NOT include DDx based on histologic, cytologic, or biopsy/pathology findings.
\\              
- The DDx list must be explicit, with a clearly defined subject disease and at least one listed differential diagnosis.
\\
- Do not infer or add any diagnoses that are not explicitly mentioned in the text.
\\
- Ensure that the subject disease name and all DDx names are written in their full, singular form without abbreviations.
\\
- If a disease name appears as an abbreviation (e.g., "BCC"), replace it with its full form only if the full form is explicitly given in the text. 
\\\\
Output formatting rules:

- If valid DDx information is found, return in the format:
'<DDx>\{Subject Disease\} : \{DDx 1\}, \{DDx 2\}, \{DDx 3\}, ...</DDx>'
\\
- Replace '\{Subject Disease\}' with the name of the disease being differentiated.
\\
- Replace '\{DDx 1\}, \{DDx 2\}, ...' with the listed differential diagnoses.
\\
- If the subject disease is missing or unclear, or if the DDx list is ambiguous or incomplete, return:
'<DDx> NOT A DDX </DDx>'
\\\\
Original text:\{text\}
\end{promptbox}
\begin{table}[H]
\centering
\small
\setlength{\tabcolsep}{4pt}
\renewcommand{\arraystretch}{1.05}
\caption{Summary of key notations}
\label{tab:notations}
\scalebox{0.9}{
\begin{tabular}{ll}
\toprule
\textbf{Symbol} & \textbf{Description} \\
\midrule
\multicolumn{2}{l}{\textit{1. Base Data Units}} \\
$I_i$ & Clinical image. \\
$d, d_i$ & Diagnosis. \\
$r, r_i$ & Textual rationale associated with a diagnostic case. \\
$t_f$ & Taxonomic information of the final diagnosis, i.e., $(\text{path}_{\mathcal{T}_{tax}}(d_f), \text{malig}(d_f))$. \\
\addlinespace[3pt]
\midrule
\multicolumn{2}{l}{\textit{2. Datasets \& Knowledge Bases}} \\
$\mathcal{D}_{case}$ & Diagnostic cases extracted from textbooks. \\
$\mathcal{S}_1,\mathcal{S}_2,\mathcal{S}_3$ & Constructed Type~1/2/3 trajectory sets. \\
$\mathcal{T}_{tax}$ & Hierarchical diagnostic taxonomy. \\
$\mathcal{G}_{DDx}$ & Differential diagnosis (DDx) graph. \\
\addlinespace[3pt]
\midrule
\multicolumn{2}{l}{\textit{3. Diagnostic Trajectory Components}} \\
$\tau^{(1)}, \tau^{(2)}, \tau^{(3)}$ & Three trajectory types (Type~1, 2, 3). \\
$d_p, r_p$ & Primary diagnosis and its rationale. \\
$d_d, r_d$ & Differential diagnosis and its rationale. \\
$\rho$ & Differential reasoning statement. \\
$d_f, t_f$ & Final diagnosis and its taxonomic information. \\
\addlinespace[3pt]
\midrule
\multicolumn{2}{l}{\textit{4. Functions \& Indices}} \\
$\text{path}_{\mathcal{T}_{tax}}(\cdot)$ & Hierarchical taxonomy path of a diagnosis. \\
$\text{malig}(\cdot)$ & Malignancy indicator (benign / malignant / precancerous in situ). \\
$\mathcal{C}$ & Candidate DDx neighbor set in \textsc{ResolveDDXNeighbor}. \\
$N$ & Number of diagnostic cases. \\
\bottomrule
\end{tabular}}
\end{table}

\section{Details of Trajectory Construction}
\label{appendix:dataset_construction}

In this section, the construction details of 3 types of trajectory dataset $\mathcal{S}_1, \mathcal{S}_2, \mathcal{S}_3$ are provided. The sample sizes for Type~1, Type~2, and Type~3 are 220, 900, and 900, respectively, totaling $K{=}2{,}020$. All trajectories are used and randomly shuffled.
\paragraph{Synthesis of Type~1 Trajectories.}
To generate Type~1 diagnostic trajectories, we synthesize direct image-to-diagnosis mappings based on the curated diagnostic case set $\mathcal{D}_{case}=\{(I_i, \;  r_i, \; d_i)\}_{i=1}^N$, where $I_i$ denotes the clinical image, $r_i$ the corresponding diagnostic rationale, and $d_i$ the confirmed diagnosis.
As summarized in \cref{alg:type1}, the procedure iterates over each record $(I_i, r_i, d_i) \in \mathcal{D}_{case}$
For each case, we retrieve the associated taxonomic information $t_f$, which includes the hierarchical path of $d_i$ in $\mathcal{T}_{tax}$, i.e., $\text{\emph{path}}_{\mathcal{T}_{tax}}(d_i)$ and its malignancy indicator $\text{\emph{malig}}(d_i) \in \{\text{\emph{benign, malignant, precancerous in situ}}\}$.
A Type~1 trajectory instance is then constructed as

$\tau^{(1)} \gets (I_i, r_i, d_i, t_f)$
representing a direct, single-step reasoning trace from the visual input to the final diagnosis.

\begin{algorithm}[H]
\caption{Synthesis of Type~1 Trajectories}
\label{alg:type1}
\begin{algorithmic}[1]
\Require Dataset $\mathcal{D}_{case}=\{(I_i, \;  r_i, \; d_i)\}_{i=1}^N$, taxonomy $\mathcal{T}_{tax}$
\Ensure Set of trajectories $\mathcal{S}_1$

\State $\mathcal{S}_1 \gets \emptyset$
\For{$(I_i,r_i,d_i)\in\mathcal{D}_{case}$}
    \State $t_f \gets (\text{\emph{path}}_{\mathcal{T}_{tax}}(d_i), \text{\emph{malig}}(d_i))$
    \State $\tau^{(1)} \gets (I_i, r_i, d_i, t_f)$
    \State $\mathcal{S}_1 \gets \mathcal{S}_1 \cup \{\tau^{(1)}\}$
\EndFor
\State \textbf{return} $\mathcal{S}_1$
\end{algorithmic}
\end{algorithm}

\paragraph{Synthesis of Type~2 Trajectories.}
Type~2 trajectories introduce differential diagnosis (DDX) reasoning while retaining the correct primary diagnosis.  
As described in \cref{alg:type2}, for each diagnostic case $(I_i, r_p, d_p) \in \mathcal{D}$, the algorithm first retrieves a clinically related differential diagnosis $d_d$ by querying the differential diagnosis graph $\mathcal{G}_{DDx}$ and the taxonomy $\mathcal{T}_{tax}$ through the \textsc{ResolveDDXNeighbor} function. If 
there is no valid neighbor, a random alternative disease is sampled from $\mathcal{D} \setminus \{d_p\}$ to ensure diversity and maintain balanced differential supervision.  
For a selected differential case $(I_j, r_d, d_d)$, an LLM comparator generates the explicit differential reasoning statement $\rho = \mathrm{LLM}(r_p, r_d, d_p)$, which highlights the discriminative features that distinguish $d_p$ from $d_d$.  
Since this trajectory represents a non-revision scenario, the final diagnosis remains unchanged ($d_f = d_p$), and its corresponding taxonomic information is denoted as $t_f = (\text{path}_{\mathcal{T}_{tax}}(d_f), \text{malig}(d_f))$. Thus, 
a complete Type~2 trajectory instance is constructed as
\[
\tau^{(2)} = (I_i, r_p, d_p, r_d, d_d, \rho, d_f, t_f),
\]
and appended to the set of trajectories $\mathcal{S}_2$.  
This process yields a collection of contrastive yet stable diagnostic examples in which differential reasoning is explicitly modeled but the final decision remains consistent with the primary diagnosis, thereby teaching the model to resist misleading differentials while maintaining diagnostic confidence.

\begin{algorithm}[H]
\caption{Synthesis of Type~2 Trajectories (No Revision)}
\label{alg:type2}
\begin{algorithmic}[1]
\Require Dataset $\mathcal{D}=\{(I_i, \;  r_i, \; d_i)\}_{i=1}^N$, taxonomy $\mathcal{T}_{tax}$, DDX graph $\mathcal{G}_{DDx}$, LLM comparator
\Ensure Set $\mathcal{S}_2$
\State $\mathcal{S}_2 \gets \emptyset$
\For{$(I_i,r_p,d_p)\in\mathcal{D}$}
    \State $d_d \gets \Call{ResolveDDXNeighbor}{d_p, \mathcal{G}_{DDx}, \mathcal{T}_{tax}, \mathcal{D}_{case}}$
    \If{$d_d = \textsc{None}$} \State $d_d \gets \Call{RandomSample}{\mathcal{D}_{case} \setminus \{d_p\}}$ \EndIf
    \State select $(I_j,r_d,d_d)\in\mathcal{D}_{case}$ with $d_d$ 
    \State $\rho \gets \mathrm{LLM}(r_p,r_d, d_p)$
    \State $d_f \gets d_p$;\quad $t_f \gets (\text{path}_{\mathcal{T}_{tax}}(d_f),\text{malig}(d_f))$
    \State $\tau^{(2)} \gets (I_i,r_p,d_p,r_d,d_d,\rho,d_f,t_f)$
    \State $\mathcal{S}_2 \gets \mathcal{S}_2 \cup \{\tau^{(2)}\}$
\EndFor
\State \textbf{return} $\mathcal{S}_2$
\end{algorithmic}
\end{algorithm}

\paragraph{Synthesis of Type~3 Trajectories.}
Type~3 trajectories extend the differential diagnosis (DDX) reasoning introduced in Type~2 by modeling \emph{reflective revision}, in which the model corrects an initially incorrect primary diagnosis after differential comparison.  
As shown in \cref{alg:type3}, the overall procedure follows the same data retrieval and pairing process as in Type~2: for each diagnostic case $(I_i, r_p, d_p) \in \mathcal{D}_{case}$, a differential counterpart $d_d$ is first identified using the \textsc{ResolveDDXNeighbor} function over the DDX graph $\mathcal{G}_{DDx}$ and taxonomy $\mathcal{T}_{tax}$.  
If no suitable neighbor exists, a random disease is sampled from $\mathcal{D}_{case} \setminus \{d_p\}$ to ensure diversity.  
The paired differential case $(I_j, r_d, d_d)$ is then used to generate an explicit reasoning statement $\rho = \mathrm{LLM}(r_p, r_d, d_d)$, emphasizing the reflective process that justifies the correction.  
In this scenario, the final diagnosis is revised to the differential one ($d_f = d_d$), and its taxonomic information is denoted as $t_f = (\text{path}_{\mathcal{T}_{tax}}(d_f), \text{malig}(d_f))$.  
A Type~3 trajectory instance is thus represented as
\[
\tau^{(3)} = (I_i, r_p, d_p, r_d, d_d, \rho, d_f, t_f),
\]
and collected into the trajectory set $\mathcal{S}_3$.  
Compared with Type~2 trajectories, which encode stable reasoning without revision, Type~3 trajectories explicitly capture the corrective dynamics of clinical reasoning, enabling the model to internalize reflective judgment and improve robustness when confronting ambiguous or misleading visual cues.

\begin{algorithm}[H]
\caption{Synthesis of Type~3 Trajectories (Reflective Revision)}
\label{alg:type3}
\begin{algorithmic}[1]
\Require Dataset $\mathcal{D}_{case}=\{(I_i, \;  r_i, \; d_i)\}_{i=1}^N$, taxonomy $\mathcal{T}_{tax}$, DDX graph $\mathcal{G}_{DDx}$, LLM comparator
\Ensure Set $\mathcal{S}_3$
\State $\mathcal{S}_3 \gets \emptyset$
\For{$(I_i,r_p,d_p)\in\mathcal{D}_{case}$}
    \State $d_d \gets \Call{ResolveDDXNeighbor}{d_p, \mathcal{G}_{DDx}, \mathcal{T}_{tax}, \mathcal{D}_{case}}$
    \If{$d_d = \textsc{None}$} \State $d_d \gets \Call{RandomSample}{\mathcal{D}_{case} \setminus \{d_p\}}$ \EndIf
    \State select $(I_j,r_d,d_d)\in\mathcal{D}_{case}$ with $d_d$
    \State $\rho \gets \mathrm{LLM}(r_p,r_d, d_d)$
    \State $d_f \gets d_d$;\quad $t_f \gets (\text{path}_{\mathcal{T}_{tax}}(d_f),\text{malig}(d_f))$
    \State $\tau^{(3)} \gets (I_i,r_p,d_p,r_d,d_d,\rho,d_f,t_f)$
    \State $\mathcal{S}_3 \gets \mathcal{S}_3 \cup \{\tau^{(3)}\}$
\EndFor
\State \textbf{return} $\mathcal{S}_3$
\end{algorithmic}
\end{algorithm}

\paragraph{Differential Diagnosis (DDX) Fallback Policy.}
To ensure reliable differential pairing even when direct graph connections are missing, we design a hierarchical DDX fallback policy, implemented in \cref{alg:resolveddx}.  
Given a diagnosis node $d$, the algorithm first queries its neighborhood in the differential diagnosis graph $\mathcal{G}_{DDx}$.  
If $d$ is not present in $\mathcal{G}_{DDx}$ or lacks adjacent nodes, its taxonomy parent in $\mathcal{T}_{tax}$ is queried as a higher-level surrogate.  
Once a set of candidate neighbors $\mathcal{C}$ is obtained, the algorithm searches for a diagnosis $c \in \mathcal{C}$ that has at least one associated case $(I_k, r_k, d_k)$ in the dataset $\mathcal{D}_{case}$.  
If such a case does not exist, the algorithm recursively examines the taxonomy children of $c$ to locate a subclass $c'$ with available samples.  
If neither direct neighbors nor fallback subclasses yield a match, the function returns \textsc{None}.

This hierarchical fallback mechanism guarantees that most differential pairs can be grounded in clinically related disease entities while preserving semantic consistency across the taxonomy and DDX graph. 
It balances coverage and precision—providing robust DDX retrieval even for underrepresented conditions—thereby preventing the sparse nodes in $\mathcal{G}_{DDx}$ from hindering the construction of reasoning-aware trajectories.

\begin{algorithm}[H]
\caption{ResolveDDXNeighbor: Find a DDX neighbor diagnosis with taxonomy fallbacks}
\label{alg:resolveddx}
\begin{algorithmic}[1]
\Require Diagnosis $d$, DDX graph $\mathcal{G}_{DDx}$, taxonomy $\mathcal{T}_{tax}$, dataset $\mathcal{D}_{case}$
\Ensure A neighbor diagnosis $\tilde d$ or \textsc{None}
\State $\mathcal{C} \gets \emptyset$
\If{$d \in \mathcal{V}(\mathcal{G}_{DDx})$ \textbf{and} $\mathrm{Adj}_{\mathcal{G}_{DDx}}(d)\neq\emptyset$}
    \State $\mathcal{C}\gets \mathrm{Adj}_{\mathcal{G}_{DDx}}(d)$
\Else
    \State $p \gets \mathrm{parent}_{\mathcal{T}_{tax}}(d)$
    \If{$p \in \mathcal{V}(\mathcal{G}_{DDx})$ \textbf{and} $\mathrm{Adj}_{\mathcal{G}_{DDx}}(p)\neq\emptyset$}
    \State $\mathcal{C}\gets \mathrm{Adj}_{\mathcal{G}_{DDx}}(p)$
    \Else
    \State \textbf{return} \textsc{None}
    \EndIf
\EndIf
\ForAll{$c \in \mathcal{C}$}
    \If{exists $(I_k,r_k,d_k)\in\mathcal{D}_{case}$ with $d_k=c$}
        \State \textbf{return} $c$
    \Else
        \State $\mathcal{S}\gets \mathrm{children}_{\mathcal{T}_{tax}}(c)$
        \If{exists $(I_k,r_k,d_k)\in\mathcal{D}_{case}$ with $d_k \in \mathcal{S}$}
            \State \textbf{return} any $c'\in \mathcal{S}$
        \EndIf
    \EndIf
\EndFor
\State \textbf{return} \textsc{None}
\end{algorithmic}
\end{algorithm}

During DDx pairing, we enforce a non-repetition constraint and increase the sampling budget until the ratio between valid DDx-based pairings and random pairings reaches approximately 1:1. This design ensures sufficient coverage when DDx candidates are exhausted as well as introducing a controlled amount of non–DDx-based reflective pairing. This design moderately encourages reflection beyond DDx-based reasoning, promoting broader and more generalizable reflective behavior.

\section{Examples of Reasoning Trajectories}
\label{appendix:examples_reasoning_trajectories}
\begin{promptbox}{An example of the type~1 trajectory}
\begin{center}
\includegraphics[width=0.25\linewidth]{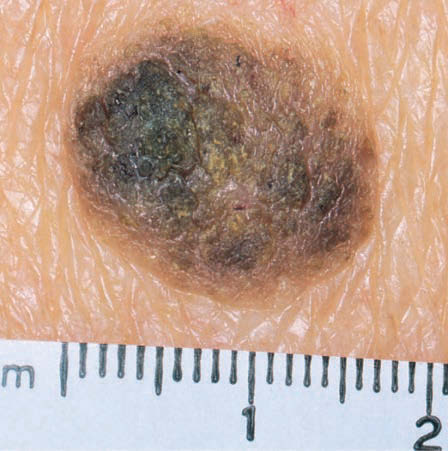}
\end{center}

\textbf{Prompt}: You are a medical vision-language assistant specializing in dermatology.
Given the dermatology image, analyze the visual findings and provide a structured diagnosis following this format:
\\\\
\texttt{<thinking>}Describe the key clinical features, visual observations, and diagnostic rationale that support the diagnosis.\texttt{</thinking>}\\
\texttt{<diagnosis>}Provide the most likely fine-grained diagnosis, and briefly classify the condition (benign or malignant) along with any relevant clinical taxonomy.\texttt{</diagnosis>}
\\\\
Ensure your response is medically accurate, concise, and strictly follows the specified format.
\rule{\linewidth}{0.4pt}

\textbf{Response}:
\texttt{<thinking>}Presence of a rough surface and a ``stuck-on'' appearance indicates seborrheic keratosis.\texttt{</thinking>}\\
\texttt{<diagnosis>}Seborrheic keratosis. It is a subtype of benign keratosis-like lesion and is classified as benign.\texttt{</diagnosis>}
\end{promptbox}

\begin{promptbox}{An example of the type~2 trajectory}
\begin{center}
\includegraphics[width=0.25\linewidth]{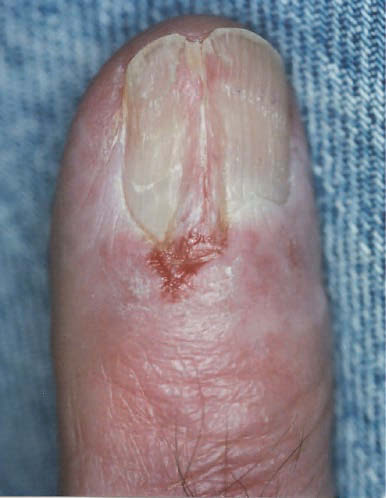}
\end{center}

\textbf{Prompt}: You are a medical vision-language assistant specializing in dermatology.
Given the dermatology image, analyze the visual findings and provide structured diagnostic reasoning following this format:
\\\\
\texttt{<thinking>}\\
Begin by describing the key clinical features and visual observations of the lesion.
Based on these features, propose the most likely diagnosis.
Then identify at least one plausible alternative diagnosis and describe its defining features.
Compare the observed lesion with these alternatives and explain step by step why the final diagnosis is more consistent with the findings.
Conclude with the single condition that best matches the clinical presentation.\\
\texttt{</thinking>}\\
\texttt{<diagnosis>}Provide only the single most likely fine-grained diagnosis. Also briefly classify the condition (benign or malignant) and indicate any relevant clinical taxonomy (e.g., subtype or disease family).\texttt{</diagnosis>}
\\\\
Ensure your response is medically accurate, concise, and strictly follows the specified format.
\rule{\linewidth}{0.4pt}

\textbf{Response}:
\texttt{<thinking>}\\
Based on the rule: Presence of an erythematous, scaly, thin plaque around the cuticular margin or a verrucous plaque in the periungual region with nail discoloration, onycholysis, or destruction of the nail plate indicates Bowen disease of the nail unit. A plausible primary diagnosis is Bowen disease of the nail unit.\\
Considering the differential diagnosis of actinic keratosis, we compare against: Presence of hypertrophic actinic keratoses on severely sun-damaged dorsal arm skin indicates actinic keratosis.\\
Here, the lesion is periungual with an erythematous, scaly thin plaque and associated nail changes (discoloration/onycholysis), consistent with Bowen disease of the nail unit. In contrast, features typical of actinic keratosis---hypertrophic lesions on severely sun-damaged dorsal arm skin---are not present. Therefore, the most likely diagnosis is Bowen disease of the nail unit.\\
\texttt{</thinking>}\\
\texttt{<diagnosis>}Bowen disease of the nail unit. It is a subtype of Bowen disease (squamous cell carcinoma in situ) and is classified as malignant.\texttt{</diagnosis>}
\end{promptbox}

\begin{promptbox}{An example of the type~3 trajectory}
\begin{center}
\includegraphics[width=0.25\linewidth]{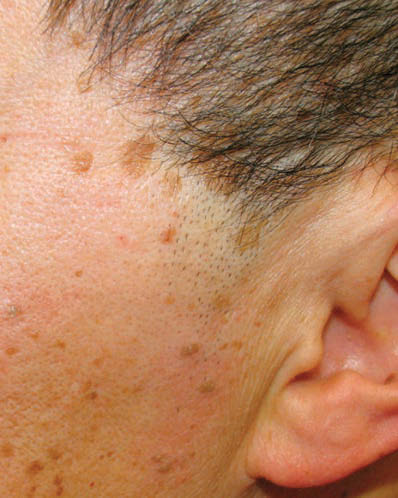}
\end{center}

\textbf{Prompt}: You are a medical vision-language assistant specializing in dermatology.
Given the dermatology image, analyze the visual findings and provide structured diagnostic reasoning following this format:
\\\\
\texttt{<thinking>}\\
Begin by describing the key clinical features and visual observations of the lesion.
Based on these features, propose the most likely diagnosis.
Then identify at least one plausible alternative diagnosis and describe its defining features.
Compare the observed lesion with these alternatives and explain step by step why the final diagnosis is more consistent with the findings.
Conclude with the single condition that best matches the clinical presentation.\\
\texttt{</thinking>}\\
\texttt{<diagnosis>}Provide only the single most likely fine-grained diagnosis. Also briefly classify the condition (benign or malignant) and indicate any relevant clinical taxonomy (e.g., subtype or disease family).\texttt{</diagnosis>}
\\\\
Ensure your response is medically accurate, concise, and strictly follows the specified format.
\rule{\linewidth}{0.4pt}

\textbf{Response}:
\texttt{<thinking>}\\
Based on the rule: Presence of multiple pigmented patches or thin papules with a reticulated pattern indicates reticulated seborrheic keratoses. An initial diagnosis could be reticulated seborrheic keratosis.\\
Considering melanoma as a differential, we compare against: Marked asymmetry, irregular borders, and multiple color variations can indicate melanoma.\\
In this case, the lesion shows irregularly distributed pigmentation with asymmetry and color variegation, which is more consistent with melanoma than a uniform reticulated pattern. The absence of characteristic multiple thin papules with a consistent reticulated network argues against reticulated seborrheic keratosis. Therefore, the final diagnosis is melanoma.\\
\texttt{</thinking>}\\
\texttt{<diagnosis>}Melanoma. It is classified as malignant.\texttt{</diagnosis>}
\end{promptbox}

\section{Experiment Implementation Details}
\label{appendix:implementation_details}
Training is performed using two NVIDIA A100 GPUs (40~GB VRAM each). We adopt Qwen2.5-VL-7B-Instruct~\citep{bai2025qwen2} as the backbone model. Both SFT and RL stages are trained with the LoRA technique~\citep{hu2022lora}, using the following hyperparameters: \texttt{lora\_r=64}, \texttt{lora\_alpha=32}, and \texttt{lora\_dropout=0.1}.  
All other hyperparameters follow the default settings of the TRL library~\citep{vonwerra2022trl}.

The SFT experiments are implemented using the TRL library~\citep{vonwerra2022trl}. Training is conducted with a total batch size of 16 and a learning rate of $3 \times 10^{-5}$ for 4 epochs.  

RL experiments are implemented and modified on the basis of the OpenR1 framework~\citep{openr1}. Training is carried out with a total batch size of 16 and a learning rate of $1 \times 10^{-5}$ for 1 epoch. The group size $K$ in GRPO is set to 4, and the sampling temperature is fixed at 1.0. The reference model is initialized from the SFT checkpoint. The maximum input image resolution is capped at $448 \times 448 = 200{,}704$ pixels.

During the evaluation, to guarantee the certainty of the result, greedy decoding is applied, which means that all the models are set as \texttt{do\_sample = False}. The maximum input image resolution is also set to $448 \times 448 = 200{,}704$ pixels.

\paragraph{Details of Baselines.}
\begin{enumerate}
    \item LLaVA-v1.6-7B~\citep{liu2024llavanext} corresponds to the Hugging Face release \emph{liuhaotian/llava-v1.6-mistral-7b}. 
    \item LLaVA-Med-7B~\citep{li2023llava} corresponds to \emph{microsoft/llava-med-v1.5-mistral-7b}.
    \item HuatuoGPT-Vision-7B~\citep{chen2024huatuogpt} corresponds to \emph{FreedomIntelligence/HuatuoGPT-Vision-7B-Qwen2.5VL}.
\end{enumerate}

In particular, due to their limited CoT reasoning and instruction-following capabilities, the prompts for LLaVA-Med-7B and SkinVL-PubMM are simplified to ensure  meaningful response generation. Detailed prompt templates are provided in Appendix~\cref{appendix:prompts}.

\section{Details of Prompts}  
\label{appendix:prompts}

\begin{promptbox}{Prompt template for SFT, type 1}
You are a medical vision-language assistant specializing in dermatology. 
Given the dermatology image, analyze the visual findings and provide a structured diagnosis following this format:
\\\\
<thinking>Describe the key clinical features, visual observations, and the diagnostic rationale that support the diagnosis.</thinking>
<diagnosis>Provide the most likely and specific fine-grained diagnosis, and briefly classify the condition (benign or malignant) along with any relevant clinical taxonomy.</diagnosis>
\\\\
Ensure your response is medically accurate, concise, and strictly follows the specified format.
\end{promptbox}
\vspace{0.8em} 
\begin{promptbox}{Prompt template for SFT, type 2 and 3}
You are a medical vision-language assistant specializing in dermatology. 
Given the dermatology image, analyze the visual findings and provide a structured diagnostic reasoning following this format:
\\\\
<thinking>
Begin by describing the characteristic clinical features and visual observations of the lesion. 
Based on these features, propose the most likely diagnosis. 
Then identify at least one plausible alternative diagnosis and describe its defining features. 
Compare the observed lesion with these alternatives and explain step by step why the final diagnosis is more consistent with the findings. 
Conclude the reasoning with the single condition that best matches the clinical presentation.
</thinking>
\\\\
<diagnosis>
Provide only the single most likely and specific fine-grained diagnosis. 
Also briefly classify the condition (benign or malignant) and indicate any relevant clinical taxonomy (e.g., subtype or disease family). 
Do not include explanations, multiple options, or extra punctuation.
</diagnosis>
\\\\
Ensure your response is medically accurate, concise, and strictly follows the specified format.
\end{promptbox}
\vspace{0.8em} 
\begin{promptbox}{Prompt template for RL training}
\textbf{Question example} = What type of abnormality is present in this image? A: actinic keratosis B: melanoma C: benign keratosis-like lesion D: squamous cell carcinoma
\rule{\linewidth}{0.4pt}
You are a medical vision-language assistant specializing in dermatology. Given the dermatology image, answer: \textbf{Question example}\\\\Provide necessary reasoning and only answer the question in the following format:\\\\<thinking>Describe the key clinical features and visual observations that support your diagnosis.</thinking>\\\\<final diagnosis>Provide only the single most likely option without reasoning. If multiple options are plausible, always choose the most fine-grained (i.e., most specific or detailed) option available among them. Also provide the lesion condition (benign, malignant or precancerous in situ). </final diagnosis>\\\\Ensure your response is medically accurate, concise, and strictly follows the specified format.
\end{promptbox}
\vspace{0.8em} 

\begin{promptbox}{Prompt template for disease diagnosis}
\textbf{Question example} = What type of abnormality is present in this image? A: melanocytic nevus B: vascular lesion C: basal cell carcinoma D: squamous cell carcinoma
\rule{\linewidth}{0.4pt}
You are a medical vision-language assistant specializing in dermatology. Given the dermatology image, answer: \textbf{Question example} \\\\Provide necessary reasoning and only answer the question in the following format:\\\\<thinking>Describe the key clinical features and visual observations that support your diagnosis.</thinking>\\\\<final diagnosis>Provide only the single most likely option (A/B/C/D) without reasoning. If multiple options are plausible, always choose the most fine-grained (i.e., most specific or detailed) option available among them.  </final diagnosis>\\\\Ensure your response is medically accurate, concise, and strictly follows the specified format.
\end{promptbox}
\vspace{0.8em} 

\begin{promptbox}{Prompt template for OOD diagnosis}

\textbf{Question example} = What specific classification does this image fall under in terms of abnormality? A: Inflammatory. B: Traumatic C: Endocrine D: Infectious
\rule{\linewidth}{0.4pt}
You are a medical vision-language assistant specializing in dermatology. Given the dermatology image, answer: \textbf{Question example}                    \\\\Provide necessary reasoning and only answer the question in the following format:\\\\                    <thinking>\\\\                        Begin by describing the characteristic clinical features and visual observations of the lesion. \\\\                        Based on these features, choose the most likely diagnosis from the options. \\\\                        Then, identify at least one plausible alternative diagnosis from other options and describe its defining features. \\\\                        Compare the observed lesion with these alternatives, and explain step by step why the final diagnosis is more consistent with the findings. \\\\                        Conclude the reasoning with the single condition that best matches the clinical presentation.\\\\                        </thinking>\\\\                    <final diagnosis>Provide only the single most likely option without reasoning.  </final diagnosis>\\\\                    Ensure your response is medically accurate, concise, and strictly follows the specified format.
\end{promptbox}
\vspace{0.8em} 
\begin{promptbox}{Prompt template for skin lesion diagnosis}
You are a medical vision-language assistant specializing in dermatology. \\\\    Given the dermatology image, answer: What type of lesion condition (benign, malignant or precancerous in situ) is present in this image? \\\\    A: benign B: malignant C: precancerous in situ\\\\    Provide necessary reasoning and only answer the question in the following format:\\\\    <thinking>Describe the key clinical features and visual observations that support your diagnosis.</thinking>\\\\    <final diagnosis>Only output one option: A, B, or C.</final diagnosis>\\\\    Ensure your response is medically accurate, concise, and strictly follows the specified format.

\end{promptbox}

\vspace{0.8em} 
\begin{promptbox}{Prompt template during evaluation for LLaVA-Med-7b}
\textbf{Question example} = What specific classification does this image fall under in terms of abnormality? A: Inflammatory. B: Traumatic C: Endocrine D: Infectious
\rule{\linewidth}{0.4pt}
You are a medical vision-language assistant specializing in dermatology. \\\\    Given the dermatology image, answer: \textbf{Question example}\\Give the correct option. Answer:

\end{promptbox}

\vspace{0.8em} 
\begin{promptbox}{Prompt template during evaluation for SkinVL-PubMM}
\textbf{Question example} = A: Inflammatory. B: Traumatic C: Endocrine D: Infectious

\rule{\linewidth}{0.4pt}
This is a skin lesion image. From the following categories: \textbf{Question example}\\Which one is the diagnosis?

\end{promptbox}

\section{Qualitative Analysis}
\label{appendix:qualitative_analysis}
We present five open-ended cases drawn from in-distribution samples and five close-ended cases from out-of-distribution (OOD) samples. \textcolor{green}{Green text} highlights clinically accurate descriptions that support the correct diagnosis or explicitly reason about the presence or absence of key features associated with competing differential diagnoses. \textcolor{red}{Red text} highlights clinically inaccurate descriptions that contradict the visual evidence in the image. Correct but diagnostically irrelevant statements, as well as mildly speculative reasoning that does not directly affect the diagnostic conclusion, are not highlighted.

Overall, in the open-ended setting, the base model tends to provide more verbose descriptions; however, it suffers in diagnostic accuracy—the most critical aspect of clinical reasoning. In contrast, \ourmethod{} demonstrates higher precision and consistency, reflecting stronger grounded diagnostic reasoning capabilities. In the close-ended setting, the base model often enumerates answer choices and bases its decision primarily on isolated visual features from the current image, without explicitly reasoning through the diagnostic criteria or characteristic features of alternative conditions, which may introduce clinical risk. \ourmethod{}, by contrast, explicitly evaluates competing diagnoses against their corresponding diagnostic rules and prototypical features, resulting in more comprehensive and reliable clinical reasoning.
Nevertheless, \ourmethod{} is not free from error: in some cases, it may overemphasize rule-based patterns or introduce speculative assumptions beyond the visual evidence, indicating that further improvements in robustness and calibration are still needed.

\begin{promptbox}{Open-ended case~1}
\textbf{Prompt} =  You are a medical vision-language assistant specializing in dermatology. Given the dermatology image, answer: What type of abnormality is present in this image? 
Describe the key clinical features and visual observations that support your diagnosis. And then the diagnosis. Also provide the lesion condition (benign, malignant or precancerous in situ).

\begin{center}
\includegraphics[width=0.25\linewidth]{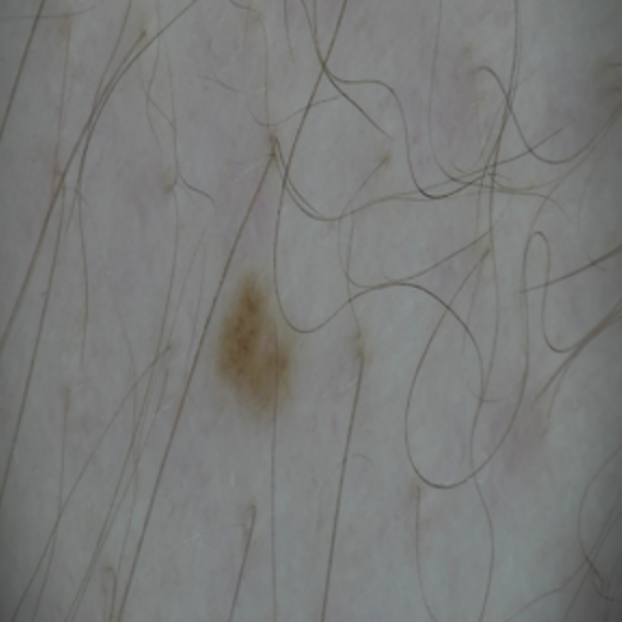}
\end{center}
\textbf{Ground truth}:melanocytic nevus, benign

\rule{\linewidth}{0.4pt}
\textbf{Response from Qwen2.5-VL-7B}:\\The image shows a skin surface with multiple fine hairs scattered across it. There is a distinct brownish patch on the skin, which appears slightly elevated and irregularly shaped.
\\\\
Key Clinical Features and Visual Observations:\\
1. **Hair Growth**: The presence of fine hairs suggests that the area might be part of the scalp or another hairy region.\\
2. **Brownish Patch**: The central brownish patch is \textcolor{red}{irregular in shape} and has a slightly raised appearance compared to the surrounding skin. This could indicate a lesion.\\
3. **Color and Texture**: The color of the patch is not uniform, suggesting variability in pigmentation, which can be a sign of melanocytic activity or other skin changes.\\
4. **\textcolor{red}{Elevation}**: The patch appears slightly elevated, which could be indicative of a lesion such as a mole, nevus, or potentially a more concerning lesion like a melanoma.\\\\

\#\#\# Diagnosis:\\
Based on the clinical features observed, the most likely diagnosis is a **pigmented lesion**. The \textcolor{red}{irregular shape}, \textcolor{red}{color variation}, and \textcolor{red}{slight elevation} suggest that this could be a **mole** (nevus) or a **\textcolor{red}{melanoma}**. However, without a biopsy and further clinical evaluation by a dermatologist, it is impossible to definitively diagnose the nature of the lesion.
\\\\
\#\#\# Lesion Condition:\\
Given the irregular shape, color variation, and slight elevation, the lesion cannot be classified as benign based on the image alone. It could be either benign (e.g., a common or atypical mole) or potentially malignant (e.g., a melanoma). Therefore, the lesion should be considered **\textcolor{red}{precancerous in situ}** until further evaluation by a dermatologist.
\\\\
\#\#\# Recommendation:\\
It is crucial to consult a dermatologist for a proper evaluation. The dermatologist may perform a biopsy to determine the exact nature of the lesion and recommend appropriate treatment if necessary. Early detection and management of skin lesions are essential for preventing potential complications.

\rule{\linewidth}{0.4pt}
\textbf{Response from \ourmethod{}}: \\The lesion exhibits the clinical features of \textcolor{green}{melanocytic nevus}, including a dark brown to black, sharply circumscribed macule with \textcolor{green}{uniform color} and \textcolor{green}{no signs of asymmetry or irregular borders}, indicating a \textcolor{green}{benign pigmented nevus} rather than melanoma. Therefore, the most likely diagnosis corresponds to "\textcolor{green}{melanocytic nevus}". The lesion is therefore classified as \textcolor{green}{benign}.
\end{promptbox}
\vspace{0.8em} 

\begin{promptbox}{Open-ended case~2}
\textbf{Prompt} =  You are a medical vision-language assistant specializing in dermatology. Given the dermatology image, answer: What type of abnormality is present in this image? 
Describe the key clinical features and visual observations that support your diagnosis. And then the diagnosis. Also provide the lesion condition (benign, malignant or precancerous in situ).

\begin{center}
\includegraphics[width=0.25\linewidth]{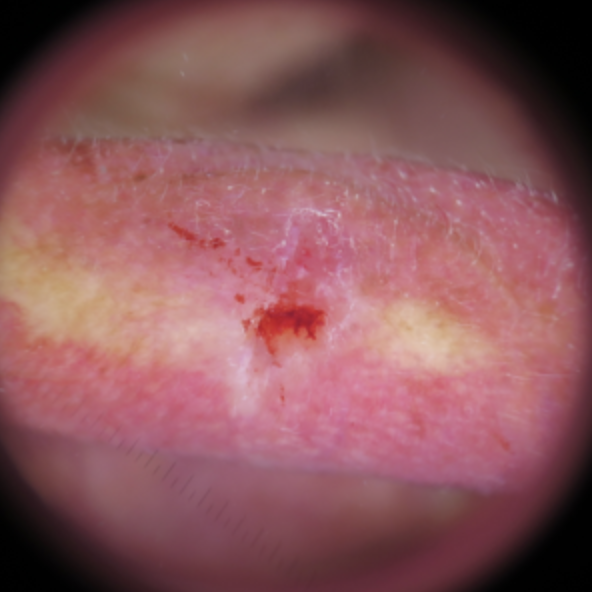}
\end{center}
\textbf{Ground truth}:basal cell carcinoma, malignant

\rule{\linewidth}{0.4pt}
\textbf{Response from Qwen2.5-VL-7B}:\\The image shows a close-up view of skin with several key clinical features:
\\\\
1. **Color Changes**: The skin appears to have \textcolor{green}{areas of redness and some yellowish discoloration}, which could indicate inflammation or infection.\\
2. **Lesion Appearance**: There is a small, irregularly shaped area of bright redness with a slightly raised border, suggesting an inflammatory process or a potential vascular issue.\\
3. **Surface Characteristics**: The surface of the lesion appears somewhat \textcolor{green}{rough and possibly ulcerated}, indicating a disruption in the normal skin barrier.
\\\\
\#\#\# Key Clinical Features Supporting Diagnosis:\\
- **Inflammatory Appearance**: The redness and surrounding changes suggest an inflammatory process.\\
- **\textcolor{green}{Ulceration}**: The presence of a small ulcer suggests a breakdown in the skin's integrity, which can be seen in various dermatological conditions such as infections, autoimmune diseases, or malignancies.
\\\\
\#\#\# Diagnosis:\\
Given the clinical features observed, the most likely diagnosis is **\textcolor{red}{inflammatory dermatitis}**. This is a broad term for a group of skin conditions characterized by inflammation. The specific type of dermatitis cannot be determined definitively from this image alone without further clinical history and examination.
\\\\
\#\#\# Lesion Condition:\\
Based on the appearance, the lesion is **\textcolor{red}{benign}**.

\rule{\linewidth}{0.4pt}
\textbf{Response from \ourmethod{}}: \\The image shows \textcolor{green}{ulceration} with a \textcolor{green}{well-defined, erythematous border} and a \textcolor{green}{central necrotic or crust-covered area}, which may appear as a shallow or deep \textcolor{green}{crater-like depression}, indicating \textcolor{green}{basal cell carcinoma}. Therefore, the most likely diagnosis corresponds to basal cell carcinoma. The lesion is generally classified as \textcolor{green}{malignant}.
\end{promptbox}
\vspace{0.8em}

\begin{promptbox}{Open-ended case~3}
\textbf{Prompt} =  You are a medical vision-language assistant specializing in dermatology. Given the dermatology image, answer: What type of abnormality is present in this image? 
Describe the key clinical features and visual observations that support your diagnosis. And then the diagnosis. Also provide the lesion condition (benign, malignant or precancerous in situ).

\begin{center}
\includegraphics[width=0.25\linewidth]{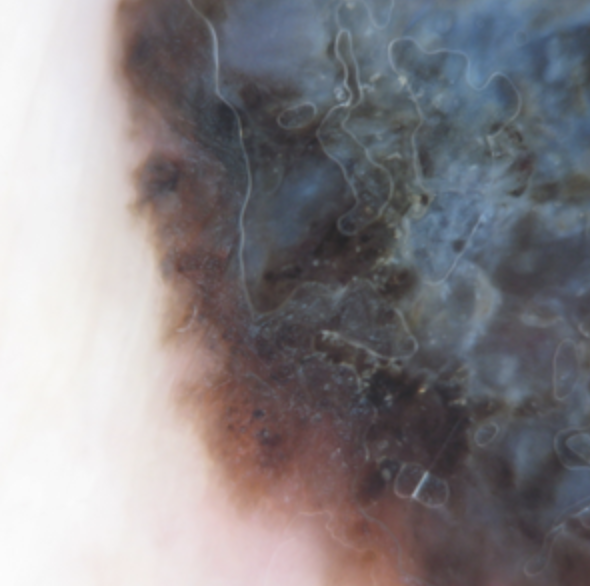}
\end{center}
\textbf{Ground truth}:melanoma, malignant

\rule{\linewidth}{0.4pt}
\textbf{Response from Qwen2.5-VL-7B}:\\
The provided image appears to be an \textcolor{red}{abstract representation} rather than a clear depiction of a skin lesion. However, based on the description of the image, it seems to show a complex pattern with irregular, wavy lines and areas of discoloration. This could potentially represent a variety of conditions depending on the context.
\\\\
\#\#\# Key Clinical Features and Visual Observations:\\
1. **Irregular Wavy Lines**: These could represent vascular structures, but without a clear view of the skin surface, it's difficult to determine their nature.\\
2. **Discoloration**: The presence of \textcolor{green}{dark and light areas} suggests possible pigmentation changes, which can be indicative of various dermatological conditions.\\
3. **Abstract Nature**: \textcolor{red}{The lack of a clear lesion} makes it challenging to provide a definitive diagnosis. This could be due to the image being abstract or a result of a specific imaging technique.
\\\\
\#\#\# Diagnosis:\\
Given the abstract nature of the image, it is not possible to provide a precise diagnosis. However, if we were to speculate based on the visual cues, the image might resemble a dermoscopic image of a mole or a skin lesion with atypical features. The irregular lines and discoloration could suggest a \textcolor{red}{premalignant} or early-stage \textcolor{green}{malignancy}, such as a \textcolor{red}{dysplastic nev}...

\rule{\linewidth}{0.4pt}
\textbf{Response from \ourmethod{}}: \\The lesion exhibits \textcolor{green}{irregular borders} with \textcolor{green}{varying shades of brown, black, and tan}, often showing \textcolor{green}{asymmetry} and multiple colors intermingled within the same lesion, which aligns with \textcolor{green}{melanoma}. Therefore, the most likely diagnosis corresponds to melanoma. The lesion is classified as \textcolor{green}{malignant}.
\end{promptbox}
\vspace{0.8em}

\begin{promptbox}{Open-ended case~4}
\textbf{Prompt} =  You are a medical vision-language assistant specializing in dermatology. Given the dermatology image, answer: What type of abnormality is present in this image? 
Describe the key clinical features and visual observations that support your diagnosis. And then the diagnosis. Also provide the lesion condition (benign, malignant or precancerous in situ).

\begin{center}
\includegraphics[width=0.25\linewidth]{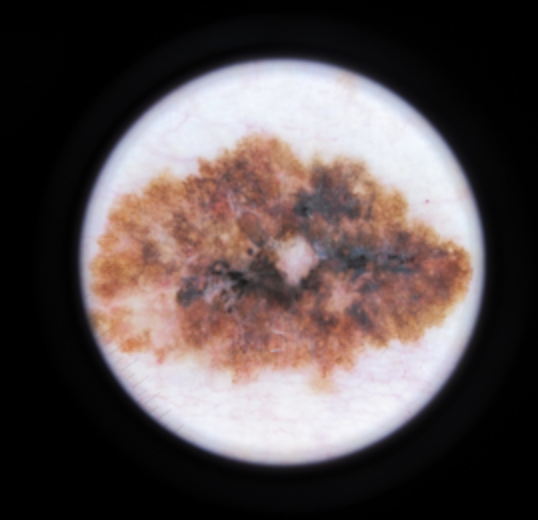}
\end{center}
\textbf{Ground truth}:melanoma, malignant

\rule{\linewidth}{0.4pt}
\textbf{Response from Qwen2.5-VL-7B}:\\
The image provided appears to be a dermoscopic view of a skin lesion. Here's an analysis based on the visual features:
\\\\
\#\#\# Key Clinical Features and Observations:\\
1. **\textcolor{green}{Asymmetry}**: The lesion does not appear symmetrical.\\
2. **\textcolor{green}{Color Variability}**: There is a mix of colors including brown, black, and reddish hues. The central area seems to have a darker, more irregular pigmentation.\\
3. **\textcolor{green}{Irregular Borders}**: The edges of the lesion are uneven and not well-defined.\\
4. **\textcolor{green}{Structureless Reddish Area}**: There is a reddish area with a structureless pattern, which can be concerning for malignancy.\\
5. **\textcolor{green}{White/Gray Areas}**: There are areas of white or gray, which could represent atrophy or depigmentation.\\
\\\\
\#\#\# Diagnosis:\\
Based on these dermoscopic features, the lesion exhibits characteristics that are often associated with \textcolor{green}{melanoma}. These include asymmetry, irregular borders, color variability, and the presence of a structureless red area, which is indicative of a junctional component.
\\\\
\#\#\# Lesion Condition:\\
Given the dermoscopic findings, this lesion is considered **\textcolor{green}{malignant}**. The presence of irregular borders, asymmetry, and the combination of colors are highly suggestive of...

\rule{\linewidth}{0.4pt}
\textbf{Response from \ourmethod{}}: \\The lesion exhibits the clinical features of \textcolor{green}{melanoma}, including \textcolor{green}{irregular borders, variable pigmentation, asymmetry, and changes in color with areas of dark brown, blue-gray, red, or gray-white}, as well as a \textcolor{green}{central tan to bluish-black macule} surrounded by a hypopigmented halo, indicating subunit melanoma. Therefore, the most likely diagnosis corresponds to \textcolor{green}{melanoma}. The lesion is \textcolor{green}{malignant}.
\end{promptbox}
\vspace{0.8em}

\begin{promptbox}{Open-ended case~5}
\textbf{Prompt} =  You are a medical vision-language assistant specializing in dermatology. Given the dermatology image, answer: What type of abnormality is present in this image? 
Describe the key clinical features and visual observations that support your diagnosis. And then the diagnosis. Also provide the lesion condition (benign, malignant or precancerous in situ).

\begin{center}
\includegraphics[width=0.25\linewidth]{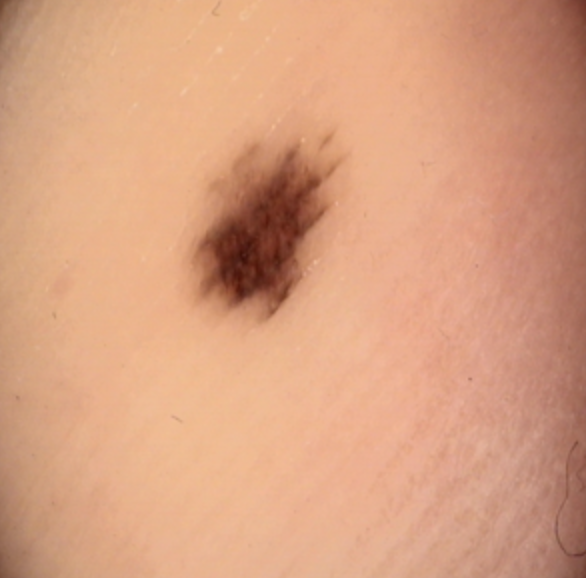}
\end{center}
\textbf{Ground truth}:melanocytic nevus, benign

\rule{\linewidth}{0.4pt}
\textbf{Response from Qwen2.5-VL-7B}:\\
The image shows a skin lesion with a dark brown to black coloration, irregular borders, and asymmetry. These are key clinical features that suggest the possibility of a melanoma.
\\\\
\#\#\# Key Clinical Features:\\
1. **Color**: The lesion has a mottled appearance with \textcolor{green}{shades of brown, black}, and possibly some lighter areas.\\
2. **Borders**: The edges of the lesion appear \textcolor{red}{irregular and not well-defined}.\\
3. **\textcolor{red}{Asymmetry}**: One side of the lesion does not match the other, which is a common feature of melanomas.\\
4. **Diameter**: While it's difficult to determine the exact size from the image alone, lesions like this often \textcolor{red}{exceed 6 mm} in diameter, which is another criterion for concern.
\\\\
\#\#\# Diagnosis:\\
Based on these clinical features, the lesion appears suspicious for a \textcolor{red}{melanoma}. However, a definitive diagnosis can only be made through a biopsy and histopathological examination by a dermatopathologist.
\\\\
\#\#\# Lesion Condition:\\
Given the clinical features observed, the lesion is considered suspicious for malignancy. It falls into the category of a "\textcolor{red}{malignant}" lesion as it exhibits characteristics that are concerning for melanoma. It is important to note that this is a preliminary assessment and a definitive diagnosis requires further evaluation by a dermat...

\rule{\linewidth}{0.4pt}
\textbf{Response from \ourmethod{}}: \\The lesion is a \textcolor{green}{uniformly dark brown, round macule with a smooth, regular border} located on the \textcolor{red}{lateral cheek of the upper eyelid}. This clinical presentation corresponds to the diagnosis of \textcolor{green}{melanocytic nevus}. Therefore, the most likely diagnosis for the lesion in the image is melanocytic nevus, and it is generally classified as \textcolor{green}{benign}.
\end{promptbox}
\vspace{0.8em}

\begin{promptbox}{Close-ended case~1}
\textbf{Prompt} =  You are a medical vision-language assistant specializing in dermatology. Given the dermatology image, answer: Is this image of a lesion indicative of being benign? A: Yes B: No.
\\
Provide necessary reasoning and only answer the question in the following format: ...\\

\begin{center}
\includegraphics[width=0.25\linewidth]{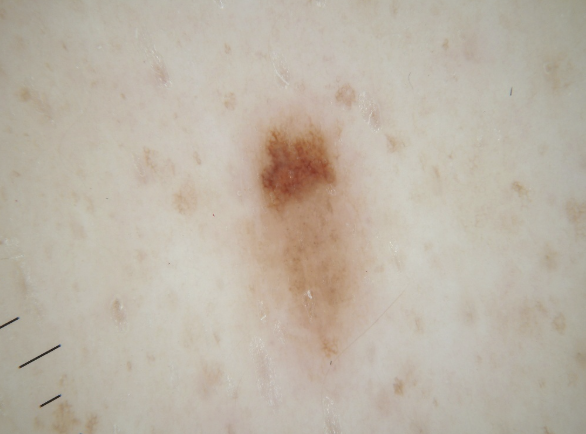}
\end{center}
\textbf{Ground truth}:A

\rule{\linewidth}{0.4pt}
\textbf{Response from Qwen2.5-VL-7B}:\\
<thinking>\\
The image shows a \textcolor{green}{brownish lesion} with \textcolor{red}{irregular borders} and \textcolor{red}{asymmetry}, which are concerning features for \textcolor{red}{malignancy}. The lesion's color, size, and shape do not align with typical benign lesions such as seborrheic keratosis or a nevus.
\\
A possible alternative diagnosis could be melanoma, characterized by irregular borders, variation in color, and asymmetry. However, the lesion in the image does not have the typical features of a melanoma, such as a "diameter greater than 6 mm" or "evolving appearance."
\\
Given the irregular borders and asymmetry, the lesion could also be a lentigo maligna, which is a premalignant lesion associated with chronic sun damage. However, lentigo maligna typically presents as a flat, brown patch with a well-defined border, which is not seen here.
\\
The lesion in the image is more suggestive of a melanoma due to its irregular borders and asymmetry, which are concerning features for malignancy.
\\
</thinking>\\
<final diagnosis>\textcolor{red}{B: No.}</final diagnosis>

\rule{\linewidth}{0.4pt}
\textbf{Response from \ourmethod{}}: \\
<thinking>\\
Based on the rule: Presence of a \textcolor{green}{brown papule} with a \textcolor{red}{blue-gray center}, beginning as a pinpoint dot in early childhood and enlarging slowly over time indicates combined common blue nevus–nevomelanocytic nevus. We can give a primary diagnosis that combined common blue nevus-melanocytic nevus. Considering the differential diagnosis for combined common blue nevus-melanocytic nevus, namely solar lentigo, we compare against the diagnostic rule for solar lentigo: Presence of light-brown macules with markedly irregular outlines on sun-exposed skin indicates solar lentigo. \\
This lesion presents as a brown papule with a blue-gray center that began as a pinpoint dot in early childhood and has enlarged slowly, which aligns with the combined common blue nevus–melanocytic nevus diagnosis. In contrast, the lesion \textcolor{green}{lacks the light-brown coloration and markedly irregular outlines} typical of solar lentigo, and it is not described as a macule on sun-exposed skin. Therefore, the most likely condition corresponds to "combined common blue nevus-melanocytic nevus". \\
</thinking>\\
<final diagnosis>\textcolor{green}{A: Yes}</final diagnosis>
\end{promptbox}
\vspace{0.8em} 

\begin{promptbox}{Close-ended case~2}
\textbf{Prompt} =  You are a medical vision-language assistant specializing in dermatology. Given the dermatology image, answer: What is the specific skin condition depicted in the image? A: Squamous cell carcinoma B: Acne C: Melanocytic nevus D: Hives
\\
Provide necessary reasoning and only answer the question in the following format: ...\\

\begin{center}
\includegraphics[width=0.25\linewidth]{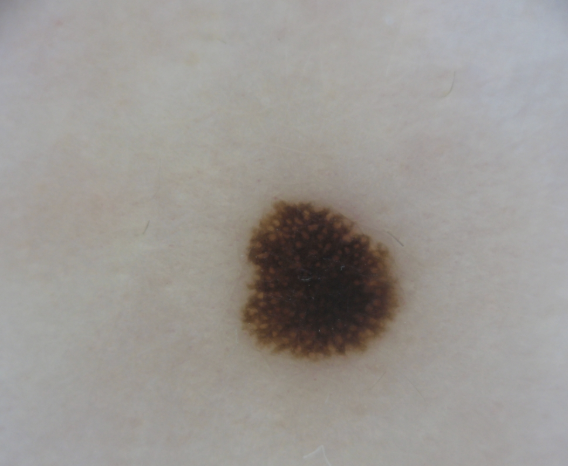}
\end{center}
\textbf{Ground truth}: C

\rule{\linewidth}{0.4pt}
\textbf{Response from Qwen2.5-VL-7B}:\\
<thinking>\\
The image shows a \textcolor{green}{well-circumscribed}, slightly raised, \textcolor{green}{brownish lesion} on the skin. It has a \textcolor{green}{uniform color and texture}, which is typical for a melanocytic nevus.
\\
A plausible alternative diagnosis could be squamous cell carcinoma (SCC), which often presents as a scaly, red, or crusty lesion with an irregular border. However, SCC typically does not have the uniform color and texture seen here.
\\
Hives (urticaria) would present as a raised, itchy area with a pale center and a red, raised border, which is not consistent with the image.
\\
Acne lesions can vary but usually do not have the uniform color and texture seen in this image.
\\
Given the uniform color, texture, and well-circumscribed appearance, the most likely diagnosis is a melanocytic nevus.\\
</thinking>\\
<final diagnosis>C: Melanocytic nevus</final diagnosis>

\rule{\linewidth}{0.4pt}
\textbf{Response from \ourmethod{}}: \\
<thinking>\\
Based on the rule: Presence of a \textcolor{green}{uniformly dark brown, round macule with a smooth, regular border} indicates a melanocytic nevus. We can give a primary diagnosis that melanocytic nevus. Considering the differential diagnosis for melanocytic nevus, namely squamous cell carcinoma, we compare against the diagnostic rule for squamous cell carcinoma: Presence of slowly enlarging, firm, skin-colored to erythematous plaques or nodules with marked hyperkeratosis on sun-exposed, heavily UV-damaged skin with multiple actinic keratoses as precursor lesions indicates cutaneous squamous cell carcinoma. \\
The lesion presents as a uniformly dark brown, round macule with a smooth, regular border, which aligns with the features of a junctional melanocytic nevus. In contrast, \textcolor{green}{there is no evidence of slowly enlarging, firm plaques or nodules with marked hyperkeratosis, nor is there mention of sun-exposed, UV-damaged skin or precursor actinic keratoses}, which are key features of squamous cell carcinoma. Therefore, the most likely condition corresponds to "melanocytic nevus". \\
</thinking>\\
    <final diagnosis>C: Melanocytic nevus</final diagnosis>
\end{promptbox}
\vspace{0.8em} 

\begin{promptbox}{Close-ended case~3}
\textbf{Prompt} =  You are a medical vision-language assistant specializing in dermatology. Given the dermatology image, answer: What is the skin abnormality present in this image? A: Rosacea B: Vitiligo C: Eczema D: Melanoma
\\
Provide necessary reasoning and only answer the question in the following format: ...\\

\begin{center}
\includegraphics[width=0.25\linewidth]{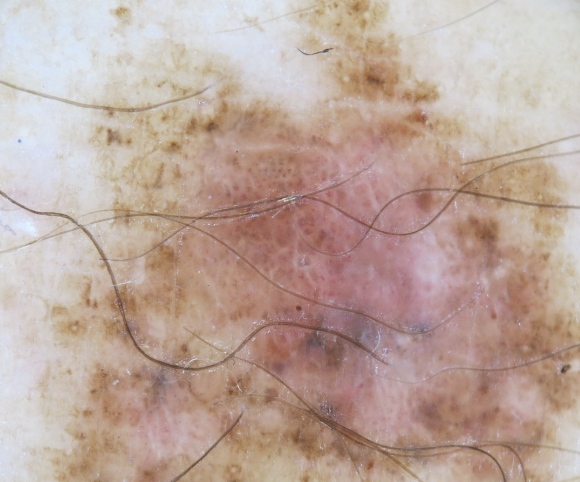}
\end{center}
\textbf{Ground truth}: D

\rule{\linewidth}{0.4pt}
\textbf{Response from Qwen2.5-VL-7B}:\\
<thinking>\\
The image shows a \textcolor{green}{red}, \textcolor{red}{raised area} with a central erythematous patch surrounded by a \textcolor{red}{ring of hyperpigmentation}. There are also some hair follicles visible within the lesion.
\\
- **Rosacea**:Typically presents as persistent facial redness, often with telangiectasias (visible blood vessels) and pustules. The central erythema and surrounding pigmentation are not typical for rosacea.
\\
- **Vitiligo**: Characterized by white patches due to loss of pigment, which is not seen here. The lesion does not have the typical depigmented appearance of vitiligo.
\\
- **Eczema**: Can present with red, itchy patches, but usually lacks the central erythema and peripheral pigmentation seen here. Eczema typically has a more diffuse distribution and may be associated with scaling or crusting.
\\
- **Melanoma**: While melanomas can vary in appearance, they often have an \textcolor{red}{irregular border, asymmetry, and a mix of colors}. This lesion \textcolor{red}{does not exhibit these characteristics}, making melanoma less likely.
\\
Given the central erythema and peripheral pigmentation, the lesion is more suggestive of a condition like \textcolor{red}{eczema} or a variant of eczema, such as nummular eczema, which can sometimes present with a ring-like pattern. However, the lack of typical eczematous features makes this less likely.
\\
The most likely diagnosis based on the provided options is:
\\
</thinking>\\
<final diagnosis>\textcolor{red}{C}</final diagnosis>

\rule{\linewidth}{0.4pt}
\textbf{Response from \ourmethod{}}: \\
<thinking>\\
Based on the rule: Presence of an \textcolor{green}{irregularly pigmented lesion with irregular borders, asymmetry, and multiple color variations such as brown, black, blue-gray, or red}, especially if located on intermittently sun-exposed skin and demonstrating structural disorganization, indicates melanoma. We can give a primary diagnosis of melanoma. To enhance the diagnostic reasoning, compare the primary diagnosis with eczema. Eczema typically presents as erythematous, pruritic patches or plaques with scaling, crusting, or lichenification, usually without marked pigment variegation or irregular borders.\\
The lesion presents as an irregularly pigmented patch with uneven borders on intermittently sun-exposed skin, showing multiple colors including brown, black, and red, which aligns with melanoma. In contrast, there is no evidence of diffuse erythematous inflammation with scaling or crusting typical of eczema. Therefore, the most likely condition corresponds to "D: Melanoma".
</thinking>\\
<final diagnosis>\textcolor{green}{D: Melanoma}</final diagnosis>
\end{promptbox}
\vspace{0.8em} 

\begin{promptbox}{Close-ended case~4}
\textbf{Prompt} =  You are a medical vision-language assistant specializing in dermatology. Given the dermatology image, answer: What is the skin abnormality present in the image? A: Rosacea B: Seborrheic keratosis C: Psoriasis D: Basal cell carcinoma
\\
Provide necessary reasoning and only answer the question in the following format: ...\\

\begin{center}
\includegraphics[width=0.25\linewidth]{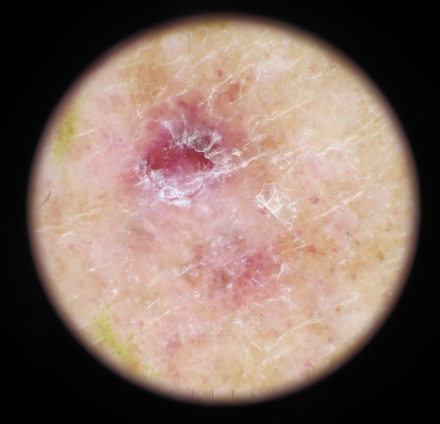}
\end{center}
\textbf{Ground truth}: D

\rule{\linewidth}{0.4pt}
\textbf{Response from Qwen2.5-VL-7B}:\\
<thinking>\\
The image shows a well-circumscribed, slightly raised lesion with a \textcolor{green}{central depression} and a slightly elevated border. The surface appears \textcolor{green}{rough} and has a \textcolor{red}{waxy appearance}. There is a central area that looks like it could be a \textcolor{red}{comedo or a plug}.
\\
- **Rosacea**: Typically presents as redness, telangiectasia, and pustules, but not usually with a central depression or comedo-like appearance.\\
- **Seborrheic Keratosis**: Usually presents as a smooth, waxy, elevated lesion with a "stuck-on" appearance, which can sometimes have a central depression or comedo-like plug.\\
- **Psoriasis**: Characterized by thick, silvery scales over erythematous plaques, which is not seen here.\\
- **Basal Cell Carcinoma**: Often presents as a pearly nodule with a rolled border, which may have a central depression, but typically lacks the waxy appearance and comedo-like plug.
\\
The lesion's waxy appearance, central depression, and comedo-like plug are more consistent with seborrheic keratosis than any of the other conditions listed.
\\
</thinking>\\
<final diagnosis>\textcolor{red}{B: Seborrheic keratosis}</final diagnosis>

\rule{\linewidth}{0.4pt}
\textbf{Response from \ourmethod{}}: \\
<thinking>\\
Based on the rule: Presence of a \textcolor{green}{pink to erythematous papule or nodule with central erosion or ulceration}, accompanied by arborizing telangiectatic vessels and a raised or rolled border, is characteristic of basal cell carcinoma. We can give a primary diagnosis of basal cell carcinoma. Considering the differential diagnosis for basal cell carcinoma, namely seborrheic keratosis, we compare against the diagnostic rule for seborrheic keratosis: Presence of a well-demarcated, hyperkeratotic, brown lesion with a “stuck-on” appearance, surface ridges, and keratin-filled cysts (horn cysts) indicates seborrheic keratosis. \\
The lesion presents as a pink erythematous papule with central erosion and telangiectatic vessels, \textcolor{green}{lacking marked hyperkeratosis appearance}, which aligns with basal cell carcinoma. In contrast, there is \textcolor{green}{no sharply demarcated brown plaque or horn cysts}, which are typical features of seborrheic keratosis. Therefore, the most likely condition corresponds to “D: Basal cell carcinoma.”
</thinking>\\
<final diagnosis>\textcolor{green}{D: Basal cell carcinoma}</final diagnosis>
\end{promptbox}
\vspace{0.8em} 

\begin{promptbox}{Close-ended case~5}
\textbf{Prompt} =  You are a medical vision-language assistant specializing in dermatology. Given the dermatology image, answer: Is this image showing a noncancerous lesion? A: No. B: Yes
\\
Provide necessary reasoning and only answer the question in the following format: ...\\

\begin{center}
\includegraphics[width=0.25\linewidth]{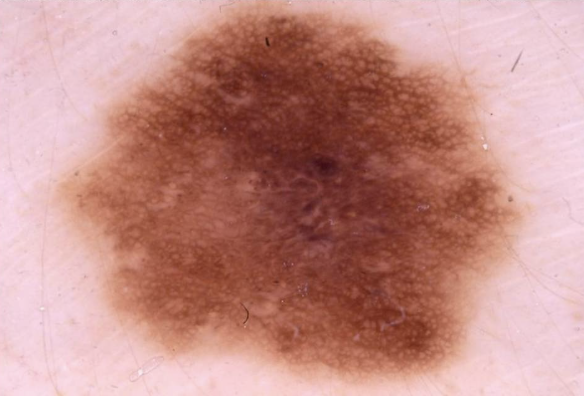}
\end{center}
\textbf{Ground truth}: B

\rule{\linewidth}{0.4pt}
\textbf{Response from Qwen2.5-VL-7B}:\\
<thinking>\\
The image shows a large, \textcolor{red}{irregularly shaped}, \textcolor{green}{brownish} lesion with a \textcolor{red}{mottled appearance}. The borders are somewhat irregular, and there is variation in color intensity within the lesion.
\\
Based on these features, the lesion is suggestive of a \textcolor{red}{melanoma}, which can present as a mole with asymmetry, irregular borders, varied colors, and an uneven surface (the ABCDEs of melanoma).
\\
A plausible alternative diagnosis could be a nevus (common or atypical mole). \textcolor{green}{Nevi typically have regular borders, uniform color, and a smooth surface}. However, some atypical nevi can have irregular borders and vary in color, making them difficult to distinguish from early melanomas.
\\
Comparing the observed lesion with these alternatives, the irregular borders, mottled appearance, and lack of symmetry are more consistent with a melanoma than a typical nevus.
\\
Conclusion: The lesion is more consistent with a melanoma.
</thinking>\\
<final diagnosis>\textcolor{red}{A: No.}</final diagnosis>

\rule{\linewidth}{0.4pt}
\textbf{Response from \ourmethod{}}: \\
<thinking> \\
The image shows a \textcolor{green}{brownish} lesion with a \textcolor{red}{uneven outline} and \textcolor{green}{subtle variation in color intensity}. The pigmentation appears predominantly brown with mild differences in shade across the lesion. While the borders are not smooth and the overall contour is \textcolor{red}{not symmetrical}. Based on these features, the lesion raises concern for \textcolor{red}{melanoma}. A plausible alternative diagnosis could be a \textcolor{green}{benign nevus}. Benign nevi often display overall symmetry, relatively uniform pigmentation, and smooth, well-defined borders. Comparing the observed lesion with these alternatives, the subtle asymmetry and irregular border contour, along with the mottled pigmentation, lean more toward melanoma rather than a clearly benign nevus. Therefore, the lesion is more consistent with melanoma rather than a definitively noncancerous lesion. 
</thinking> \\
<final diagnosis>\textcolor{red}{A: No.}</final diagnosis>

\end{promptbox}

\section{Details of granularity-wise reward $R_{\text{gran}}$}  
\label{app:granularity}

\begin{figure*}[htbp]
\centering
\includegraphics[width=0.8\textwidth]{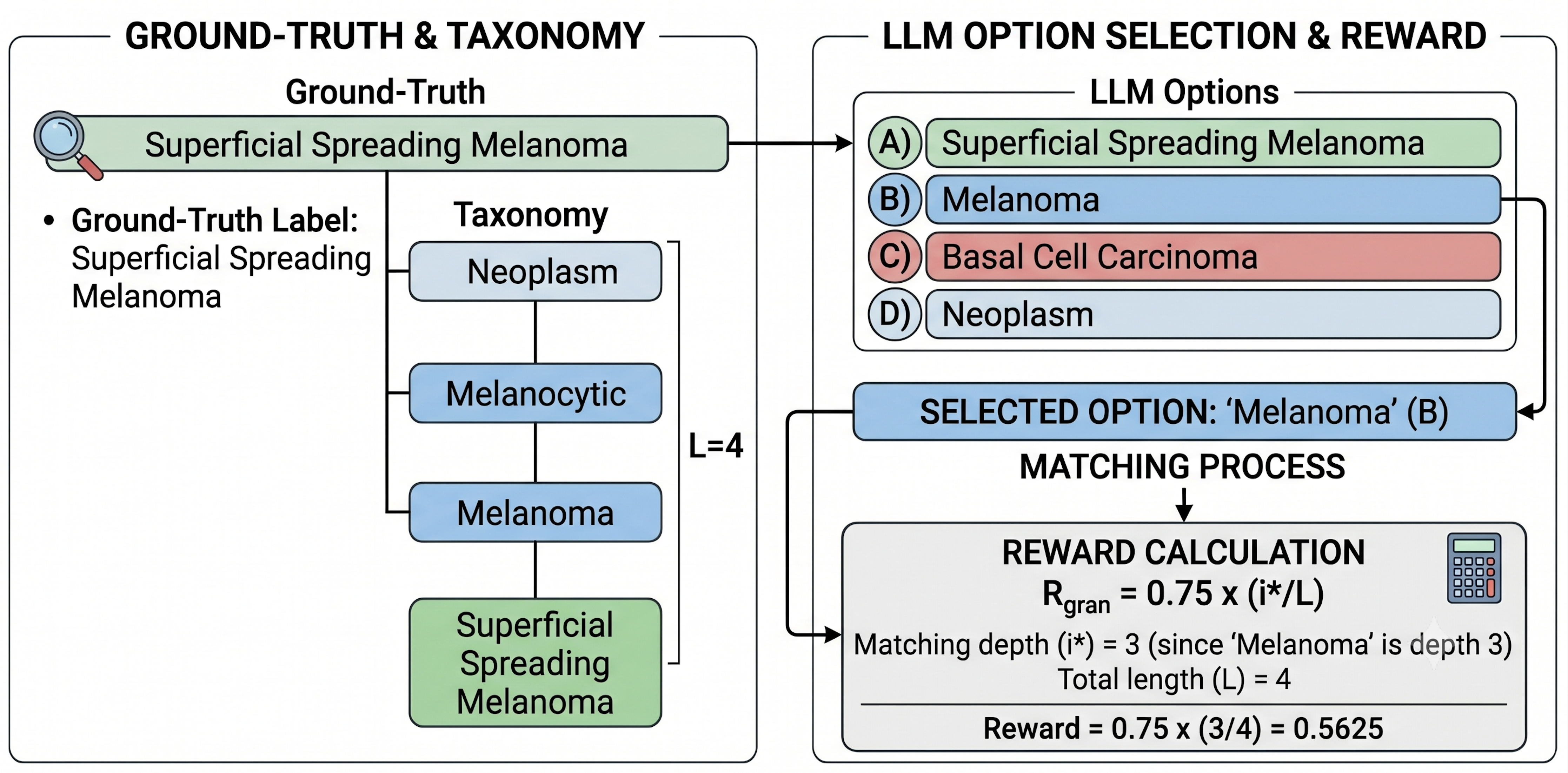}
\caption{An example of the granularity-wise reward calculation. Green represents the fine-grained correct answer, blue denotes coarse-grained correct answers, and red indicates incorrect options.}
\label{fig:hierarchial_example}
\end{figure*}

During scoring, given a ground-truth label, its full taxonomy path $\mathcal{P}$ is retrieved. $R_{\text{gran}}$ (\cref{eq:granularity}) uses exact option matching: if the selected option $\hat{\ell}$ matches any node on $\mathcal{P}$ that appears among the current options, the reward is $0.75 \times i^*/L$ ($i^*$: matched depth, $L$: path length); otherwise 0. An example is shown in \cref{fig:hierarchial_example}: For ground-truth ``Superficial Spreading Melanoma'' with path Neoplasm $\to$ Melanocytic $\to$ Melanoma $\to$ Superficial Spreading Melanoma ($L{=}4$), if all four nodes appear as options and the model selects ``Melanoma'' ($i^*{=}3$), the reward is $0.75 \times 3/4{=}0.5625$. We will add more details and an illustrative figure in the revision.

\section{Bias Evaluation Across Diverse Populations}  
\label{appendix:Fitzpatrick}

\begin{table}[htbp!]
\centering
\caption{Performance across Fitzpatrick skin types.}
\begin{tabular}{c|cccccc}
 Skin Type & I & II & III & IV & V & VI \\
\hline
Skin-R1 & 0.5254 & 0.4634 & 0.5494 & 0.5415 & 0.4917 & 0.4828 \\
\end{tabular}

\label{tab:bias_analysis}
\end{table}

To evaluate potential demographic bias in model performance, we analyze results across different skin tone groups. Fitzpatrick skin types are a dermatological classification that categorizes human skin into six types (I–VI) based on how it responds to ultraviolet (UV) exposure, particularly its tendency to burn or tan. 
On a subset of Fitzpatrick-17k~\citep{groh2021evaluating}, we observe slight performance drops on Skin Types II, V, and VI (Table~\ref{tab:bias_analysis}), suggesting potential sensitivity to population distribution. Future work will explore fairness-aware training and domain adaptation to improve robustness across diverse demographic groups.

\section{Data Availability.}
Due to copyright restrictions, we released the structured reasoning annotations and metadata of \textbf{SkinRationale}, without redistributing copyrighted images or text. 
In addition, we released the weight and code of \ourmethod{}.

\end{document}